\documentclass{article}

 \usepackage[main, final]{neurips_2026}

\usepackage[utf8]{inputenc} 
\usepackage[T1]{fontenc}    
\usepackage{hyperref}       
\usepackage{url}            
\usepackage{booktabs}       
\usepackage{amsfonts}       
\usepackage{nicefrac}       
\usepackage{microtype}      
\usepackage{xcolor}         

\usepackage{wrapfig}
\usepackage{comment}
\usepackage{nicefrac}       
\usepackage{microtype}      
\usepackage{xcolor}         
\usepackage{colortbl} 
\usepackage{subcaption}
\usepackage{multirow}
\usepackage{verbatim}
\usepackage{diagbox}
\usepackage{bm}
\usepackage{makecell}
\usepackage{graphicx}
\usepackage{amsmath}
\usepackage[english]{babel}
\usepackage{mathtools}
\usepackage[ruled,vlined]{algorithm2e}
\usepackage[utf8]{inputenc}
\usepackage{csquotes}
\usepackage{enumitem}
\usepackage{soul}
\usepackage{lipsum} 
\usepackage{booktabs} 
\usepackage{caption} 
\usepackage{rotating} 
\usepackage{multirow} 
\usepackage{lipsum} 
\usepackage{colortbl}
\usepackage{arydshln}
\usepackage{amsthm}
\usepackage{amssymb}
\usepackage{bbm}
\usepackage{hhline}   
\usepackage{makecell} 
\usepackage{xcolor}
\definecolor{orange}{HTML}{D55E00}
\definecolor{blue}{HTML}{0072B2}
\definecolor{green}{HTML}{009E73}
\definecolor{table-yellow}{HTML}{FAF4C0}
\definecolor{lime}{RGB}{204, 255, 102}
\definecolor{gray}{HTML}{EAEAEA}
\sethlcolor{lime}

\title{Logit-Conditioned Diffusion Decoding \\ for Frozen Discrete-Token VLMs}

\author{%
  \textbf{Ji Woo Hong}$^{1}$\thanks{This work was done while Ji Woo Hong was an intern at MSRA.} \quad
  \textbf{Gwanhyeong Koo}$^{1}$ \quad
  \textbf{Hee Suk Yoon}$^{1}$ \quad
  \textbf{Eunseop Yoon}$^{1}$ \\
  \textbf{SooHwan Eom}$^{1}$ \quad
  \textbf{Qi Dai}$^{2}$ \quad
  \textbf{Chong Luo}$^{2}$\thanks{Corresponding Authors.} \quad
  \textbf{Chang D. Yoo}$^{1}$\footnotemark[2] \\
  $^{1}$Korea Advanced Institute of Science and Technology (KAIST) \\
  $^{2}$Microsoft Research Asia (MSRA) \\
  \texttt{\{jiwoohong93,cd\_yoo\}@kaist.ac.kr} \\
  \texttt{\{cluo\}@microsoft.com} \\
}

\begin{document}

\maketitle

\begin{abstract}
Pre-trained discrete-token vision-language models (VLMs) generate
images by sampling sequences of codebook indices, but their visual fidelity is
bounded by the VQ-VAE decoding stage. Prior work addresses this
bottleneck by replacing the discrete tokenizer with continuous or
hybrid visual token representations, or by integrating diffusion into
the generation pipeline; both routes require foundation-scale
retraining or realignment of VLM, which is undesirable when the VLM's
broader capabilities should be preserved or when retraining is infeasible. 
We show that such retraining is unnecessary: before each token is sampled, discrete-token VLMs already compute a full logit distribution over the codebook, but standard decoding
discards it after selecting a single index. Under Otsu's threshold,
the pre-sampling logits of capable VLMs assign statistically meaningful probability
to at least two codebook entries per token, and recovering this
distribution can substantially close the fidelity gap. We propose
DCDD, a post-hoc framework that conditions a diffusion decoder on this
discarded distribution while leaving the VLM frozen. At inference,
DCDD converts the VLM's pre-sampling logits into distribution-weighted
code vectors; during training, it uses VQ-VAE encoder-derived proxy
logits calibrated to match the distributional statistics of the VLM's inference-time logits.
Our diffusion decoder then maps these representations to high-fidelity images as an
optional alternative to the native VQ-VAE decoder. Trained on
ImageNet-1K for 50K steps, DCDD reduces reconstruction FID by 74\% and Janus-Pro 7B's generation FID on MJHQ-30K by
34\%, while semantic alignment across GenEval, DPG-Bench, and WISE remain
consistent. 
Controlled comparisons with diffusion models from the same backbone
family, applied either as an image-to-image refiner or as an
identically trained decoder on the sampled codes, confirm that the gain
stems from conditioning on the VLM's pre-sampling logit distribution
rather than from the diffusion model itself.
Code is available at \url{https://github.com/jiwoohong93/dcdd_code}.
\end{abstract}

\section{Introduction}
\label{sec:intro}
\begin{figure}[t]
    \centering    
    \includegraphics[width=0.9\textwidth]{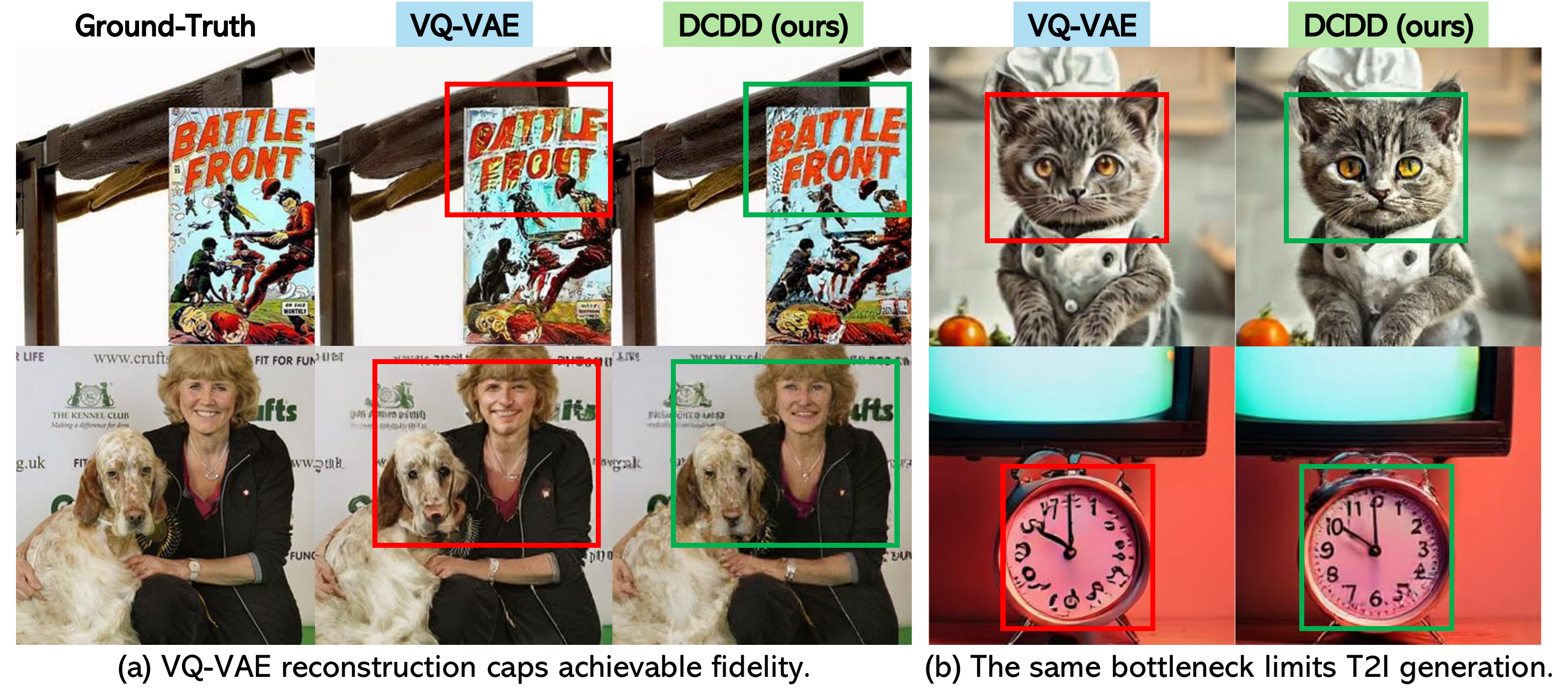}
\caption{\textbf{The VQ-VAE decoder is the fidelity bottleneck in
discrete-token VLM image generation.}
\textit{(a) Reconstruction:} even when the input image is available
and the VLM is bypassed entirely, the VQ-VAE encode--decode pipeline
already loses fine structure relative to the ground truth (book cover
text, facial details). \textit{(b) Text-to-image generation:} the
same bottleneck propagates to the deployed pipeline, where the native
VQ-VAE decoder produces distorted local geometry
on the VLM's predicted tokens. DCDD replaces only the decoding stage
and recovers fine detail with the VLM left frozen.}
   \label{fig:problem}
    \vskip -0.28in
\end{figure}

Autoregressive vision-language models (VLMs)
\cite{ramesh2021zeroshot, yu2022scaling, chen2025janus, sun2024autoregressive}
generate images by predicting discrete tokens from a learned VQ-VAE
codebook \cite{van2017neural, razavi2019generating, esser2021taming}.
Although these models exhibit strong semantic capabilities, their
images often contain distortions in local structure and geometry
(Figure~\ref{fig:problem}); as VLMs scale, the bottleneck increasingly
lies in the decoding stage rather than the transformer itself.

\begin{figure}[t]
    \centering    
    \includegraphics[width=\textwidth]{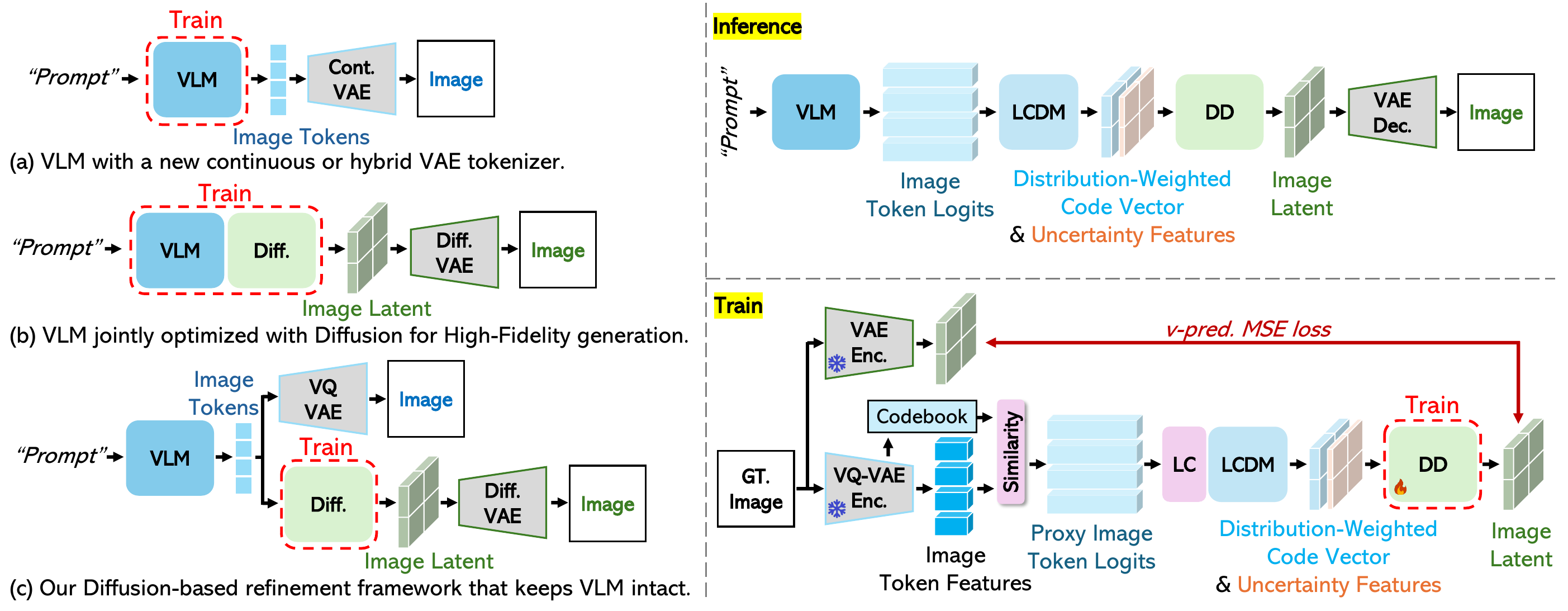}
\caption{
\textbf{\textit{Left}: Strategies for improving visual fidelity in
pre-trained discrete-token VLM image generation, characterized by
which components are trained.}
\textbf{\textit{(a) Tokenizer replacement:}} the VLM is retrained on
foundation-scale T2I data to accommodate a new continuous or hybrid
tokenizer.
\textbf{\textit{(b) Joint optimization with diffusion:}} the VLM and
a diffusion model are jointly trained, making diffusion a required
part of generation.
\textbf{\textit{(c) DCDD (ours):}} only the diffusion decoder is
trained; the VLM is left frozen, and the native VQ-VAE decoding path
remains available.
\textbf{\textit{Right}: DCDD framework.}
At inference, LCDM converts the VLM's image-token logits into conditioning for the Diffusion Decoder. At training, proxy logits from the VQ-VAE encoder are aligned to the VLM's
inference-time statistics by LC, a gradient-free transform whose
parameters are estimated beforehand.}
   \label{fig:overview}
    \vskip -0.2in
\end{figure}

Existing approaches to this fidelity bottleneck modify the system in one of two ways. 
One route replaces the original VQ-VAE tokenizer with a higher-capacity
continuous or hybrid visual token representation
(Figure~\ref{fig:overview}, \textit{left}(a)); another integrates
diffusion directly into the generation pipeline
(Figure~\ref{fig:overview}, \textit{left}(b)). In both cases, the
generation is rebuilt around a new token, so the VLM must be retrained or realigned at foundation-scale.
Both options are undesirable when the VLM's broader capabilities should be
preserved, or when retraining is infeasible. We therefore ask: can the
visual fidelity of a discrete-token VLM be improved post-hoc, with the
VLM left untouched?

Our starting point is that discrete-token VLMs already compute, but
discard, the information needed to answer this question. 
At each spatial position, the VLM produces logits over all codebook
tokens, and standard decoding reduces this distribution to a single
token index by argmax.
The full distribution, however, is not one-hot: under Otsu's
variance-maximizing threshold~\cite{otsu1979threshold}, the logits of
Janus-Pro 7B~\cite{chen2025janus}, a publicly released discrete-token
AR VLM, assign statistically meaningful probability to at least two
codebook entries per token (Table~\ref{tab:otsu}), with top-1
probability 0.22 and top-2 probability 0.10. Recovering this
distributional information can substantially close the fidelity gap.

Building on this observation, we introduce \textbf{DCDD
(Distribution-Conditioned Diffusion Decoder)}, a post-hoc framework
that conditions a diffusion decoder on the VLM's \textit{logit distribution}
rather than solely on the sampled token. DCDD comprises three components
(Figure~\ref{fig:overview}, \textit{right}). \textit{Logit-to-Code Distributional Mapping
(LCDM)} converts each token's logit distribution into continuous
distribution-weighted code vectors with complementary uncertainty
features. \textit{Logit Calibration (LC)} is a gradient-free procedure that
aligns the VQ-VAE encoder-derived proxy logits used during training
with the distribution statistics of the VLM's inference-time logits, so
that the decoder sees matched conditioning at training and inference.
\textit{The Diffusion Decoder (DD)}, finetuned from FLUX.1-Fill, maps these
representations to high-fidelity images and serves as an optional
alternative to the VQ-VAE decoder, with the VLM left frozen.

DCDD is scoped to fidelity: the diffusion
decoder is conditioned only on VLM-derived token distributions and
excludes direct text conditioning, so it follows the VLM's original
token belief rather than reinterpreting the scene. 
We primarily evaluate DCDD on Janus-Pro 7B~\cite{chen2025janus}, one of the most capable publicly released discrete-token VLMs. We also demonstrate that the decoder transfers to Janus-Pro 1B without retraining, as both models share the same tokenizer. Finally, we use LlamaGen-XL~\cite{sun2024autoregressive} to characterize when post-hoc decoding is effective.
Trained on ImageNet-1K alone, DCDD substantially improves both
reconstruction and generation fidelity while keeping semantic
consistency stable. Comparisons with diffusion models from the same
backbone family as DCDD's decoder, applied either as an image-to-image
refiner (FLUX.1 Kontext~\cite{labs2025flux}) or as an identically
trained decoder on the sampled codes, further show that DCDD's gain
stems from conditioning on the VLM's pre-sampling logit distribution
rather than from the diffusion model itself.

Our contributions are: (i) We identify and quantify a previously underused signal in discrete-token VLMs: the pre-argmax logit distribution carries rank-2 or higher codebook information that
standard decoding discards. (ii) We propose DCDD, a post-hoc framework that exploits this signal as a diffusion conditioning input, requires no VLM modification, and trains on ImageNet-1K alone. (iii) We provide controlled evidence that the gain originates from logit-level conditioning rather than from the diffusion prior or image-to-image refinement, that semantics are preserved, and that the method's effectiveness is governed by upstream token quality and logit structure.

\section{Related Work}
\label{sec:related_work}

Approaches to improving the visual fidelity of discrete-token VLM image generation fall along three axes characterized by which components are trained (Figure~\ref{fig:overview}, \textit{left}).

\textbf{Tokenizer replacement.}
One line of work replaces the discrete VQ-VAE with continuous or hybrid tokenization to reduce quantization artifacts and preserve fine-grained details. TokenBridge~\cite{wang2025bridging} and MAR~\cite{li2024autoregressive} decouple quantization from training or
eliminate the discrete codebook entirely. HART~\cite{tang2024hart} and
DC-AR~\cite{wu2025dc} adopt hybrid schemes in which discrete tokens encode global structure while continuous residuals model fine details.
These approaches achieve low reconstruction FID but require foundation-scale text-to-image retraining of the VLM.

\textbf{Joint optimization with diffusion.}
A second line integrates diffusion directly into the autoregressive generation pipeline. X-Omni~\cite{geng2025x} aligns AR generation with diffusion-based decoding through reinforcement learning. UniFusion~\cite{li2025unifusion} treats a VLM as a unified encoder that conditions a diffusion model. Show-o~\cite{xie2024show} and ILLUME+~\cite{huang2025illume+} unify multimodal understanding and generation in architectures that combine autoregressive modeling with diffusion-based decoding. These methods attain strong fidelity, but diffusion becomes a required part of generation: the VLM is jointly trained or realigned with the diffusion module, the lightweight VQ-VAE pathway is forfeited, and significant additional T2I-scale optimization is needed to preserve multimodal capabilities.

\textbf{Post-hoc decoding.}
A third axis, the one occupied by our DCDD, leaves the VLM frozen and
modifies only the decoding stage (Figure~\ref{fig:overview}, \textit{left}(c)). 
Prior work on this axis has been limited: text-conditioned image-to-image refiners~\cite{song2021score, meng2022sdedit, brooks2023instructpix2pix} can be applied to VQ-VAE outputs without retraining the VLM, but because they are guided by their own pre-trained prior, they are prone to semantic drift from the VLM's intent.
DCDD addresses this by conditioning the diffusion decoder directly on the VLM's pre-argmax logit distribution rather than on already-decoded images, so that the VLM's predictive belief guides the decoding (Section~\ref{sec:llcvspsr}). 
Soft targets~\cite{hinton2015distilling} and temperature or
label-smoothing calibration~\cite{guo2017calibration,muller2019label}
are established tools; DCDD's contribution is to identify the
pre-sampling visual-token distribution of a frozen VLM as a
recoverable signal, and to make it usable for decoder training 
through calibrated encoder proxies, without running the VLM during training.
DCDD shows that a substantial fraction of the fidelity gap can be closed with decoder-only training on ImageNet-1K, without modifying the underlying VLM.

The three axes correspond to different deployment regimes. Tokenizer
replacement and joint optimization with diffusion rebuild the generative system around a new decoding mechanism, which is appropriate when the VLM is being trained from scratch. Post-hoc decoding is appropriate when a pre-trained VLM is fixed by an upstream provider, when its broader capabilities should be preserved, or when foundation-scale retraining is not available. DCDD targets this third regime, showing that meaningful fidelity gains are achievable without foundation-scale retraining.

\section{Conceptual Motivation}
\label{sec:motivation}

DCDD rests on two empirical observations about the logit distributions involved in discrete-token VLM image generation: VLM and VQ-VAE encoder logits share a structural codebook geometry, but differ sharply in their statistical concentration. We make these observations precise here, as they directly motivate the two non-trivial design choices in Section~\ref{sec:method}: using continuous logit distributions as conditioning (LCDM) and calibrating proxy logits before training (LC).

\begin{wraptable}{r}{0.60\textwidth}
    \vskip -0.15in
    \centering
    \small
    \setlength{\tabcolsep}{4pt}
    \renewcommand{\arraystretch}{0.95}
    \caption{\textbf{Token-probability and Otsu thresholding
statistics for Janus-Pro 7B.} VLM image-token logits versus VQ-VAE
encoder proxy logits. The two distributions live in the same codebook
space but differ by orders of magnitude in concentration: the VLM
commits mass to roughly 2 entries per token, while the raw encoder
spreads it over thousands. The third column shows the encoder proxy
logits after Logit Calibration (Section~\ref{sec:lc}), which restores
sharpness comparable to the VLM.}
\begin{tabular}{lccc}
\Xhline{4\arrayrulewidth}
\rowcolor{table-yellow!60}
\textbf{Statistic} &
\textbf{VLM} &
\textbf{VQ-VAE} &
\textbf{VQ-VAE (LC)} \\
\midrule
\multicolumn{4}{l}{\textbf{\textit{Probability Distribution Statistics}}} \\
\midrule
Top-1 Probability      & 0.221627 & 0.000151 & 0.222283 \\
Top-2 Probability      & 0.100532 & 0.000145 & 0.064229 \\
Norm. Entropy     & 0.415303 & 0.993630 & 0.424155 \\
Tail Entropy           & 0.560794 & 0.993661 & 0.520536 \\
\midrule
\multicolumn{4}{l}{\textbf{\textit{Otsu Thresholding Statistics}}} \\
\midrule
Threshold (prob.)      & 6.24$\times10^{-2}$ & 6.66$\times10^{-5}$ & 6.38$\times10^{-2}$ \\
Threshold Rank         & 2        & 5767     & 2 \\
Head Mass              & 0.322160 & 0.492722 & 0.286512 \\
\Xhline{4\arrayrulewidth}
\end{tabular}
    \label{tab:otsu}
    \vskip -0.05in
\end{wraptable}

\textbf{Logits over a shared codebook.}
At each generation step, the VLM transformer projects its hidden state
into a vector of logits $l_t \in \mathbb{R}^K$, with one entry for each
VQ-VAE codebook index. A discrete token index is sampled from this
distribution, and only the selected index is passed to the VQ-VAE
decoder. The VQ-VAE encoder, in turn, produces a latent feature
$f_i \in \mathbb{R}^D$ at every spatial position and assigns it to the
nearest codebook entry by L2 distance~\cite{linde1980algorithm}. Thus, the VLM and the encoder
operate over the same codebook-index space, although they arrive at it
through different mechanisms: the VLM predicts a distribution over
codebook indices, while the encoder assigns image-patch features to
entries in the tokenizer latent space. We exploit this shared index
space by converting the encoder's assignment scores into a continuous
distribution over the codebook via cosine similarity
(Section~\ref{sec:lc}), yielding proxy logits that can be used during
training in place of the VLM's inference-time logits.

\textbf{The two distributions are statistically incompatible.}
Although the VLM and encoder operate over the same codebook-index
space, their probability distributions concentrate mass very
differently. Table~\ref{tab:otsu} reports token-probability and Otsu
thresholding statistics for Janus-Pro 7B's image-token logits and the
corresponding VQ-VAE encoder proxy logits. The VLM places concentrated
mass on a small number of codebook entries (top-1 0.22, normalized
entropy 0.42, Otsu threshold rank 2), reflecting a selective belief over
plausible visual tokens. In contrast, the raw encoder proxy
distributions are nearly uniform (top-1 $1.5 \times 10^{-4}$,
normalized entropy 0.99, Otsu threshold rank 5,767), spreading mass over
thousands of codebook entries.


This flatness is not merely a numerical artifact; it follows from using
bounded cosine similarities over a large codebook. The proxy logits
$s_{i,k}=\cos(f_i,e_k)$ lie in $[-1,1]$, and with
$K=16{,}384$ codebook entries, a top entry with cosine near $1$ and
most remaining entries near $0$ yields a softmax probability of only
approximately
$\frac{e^1}{e^1+(K-1)e^0} \approx 1.6 \times 10^{-4}$, while the
uniform probability is $1/K \approx 6.1 \times 10^{-5}$. Even under
more favorable score separation, the top probability remains extremely
small because of the large vocabulary size. Consequently, the raw
encoder distribution is only marginally sharper than uniform and does
not match the selective probability structure of the VLM logits.

\textbf{Implications.}
Two consequences follow. 
First, reducing each distribution to a single sampled token index
discards rank-2 or higher information that the VLM has already computed
(Table~\ref{tab:otsu})~\cite{hinton2015distilling}. Conditioning on a distribution-aware
representation, rather than only the sampled index, provides a compact
way to retain this additional codebook information. This motivates LCDM
(Section~\ref{sec:lcdm}). Second, proxy logits cannot be used directly
during training, because their high-entropy, near-uniform shape differs
sharply from the inference-time VLM logits that the diffusion decoder
will receive at deployment. Bridging this gap requires aligning the
encoder proxy distribution with the VLM's statistical regime, which
motivates Logit Calibration (Section~\ref{sec:lc}). As shown in the
third column of Table~\ref{tab:otsu}, Logit Calibration substantially
closes this gap, restoring the encoder's Otsu rank to 2 and bringing
its top-1 probability and head mass close to those of the VLM by
matching a small set of distributional statistics.

\section{Method}
\label{sec:method}

We now describe the three components of the DCDD framework:
Logit-to-Code Distributional Mapping (LCDM, Section~\ref{sec:lcdm}),
Logit Calibration (LC, Section~\ref{sec:lc}), and the Diffusion
decoder (DD, Section~\ref{sec:dd}) built on the denoising generative
modeling paradigm~\cite{ho2020denoising,nichol2021improved}. An overview is shown in
Figure~\ref{fig:overview}, \textit{right}. The detailed formulation,
including the full set of equations, training objective, and
architectural details, is provided in Appendix~\ref{sec:extended_method}.

\subsection{Logit-to-Code Distributional Mapping (LCDM)}
\label{sec:lcdm}
LCDM converts the VLM's per-token logits into the continuous
representation that conditions the diffusion decoder. Given
\textit{image-token logits} $l_i \in \mathbb{R}^K$ at spatial position
$i$, we form the probability distribution $p_{i,k} =
\mathrm{softmax}(l_i)_k$ over codebook entries and compute the
\textit{distribution-weighted code vector} $v_i = \sum_{k=1}^{K}
p_{i,k} e_k \in \mathbb{R}^D$, which is the expected codebook
embedding under $p_i$. 
As established in Section~\ref{sec:motivation},
this provides a compact distribution-aware alternative to the
standard argmax~\cite{hinton2015distilling}: it retains the rank-2 or
higher information that the VLM places on multiple codebook entries
per token, rather than collapsing each token to a single index.

The expected embedding alone, however, hides the shape of $p_i$. Two
tokens with very different sharpness can produce a similar $v_i$. To
preserve this distinction, we additionally extract four scalar
\textit{uncertainty features} from $p_i$: the maximum probability,
the top-1/top-2 confidence margin, an entropy-adaptive top-$K_u$
probability mass, and the normalized tail entropy. These are
concatenated into a per-token uncertainty vector $u_i \in
\mathbb{R}^4$. Together, $(v_i, u_i)$ form the conditioning input to
the decoder. Full definitions are in
Appendix~\ref{sec:extended_lcdm}.

\subsection{Logit Calibration (LC)}
\label{sec:lc}
Ideally, the diffusion decoder would be trained on logits produced by
the VLM itself, so that its training-time inputs match what it will
see at inference. This is impractical at scale: producing VLM logits
for every training image requires running the large VLM on each
example, which dominates the cost of training. We instead obtain
\textit{proxy logits} from the VQ-VAE encoder: for a ground-truth
image, we compute cosine similarity $s_{i,k} = \cos(f_i, e_k)$
between each image-token feature and every codebook entry, and apply
the same LCDM transform.

As Section~\ref{sec:motivation} establishes, however, these proxy logits
are statistically incompatible with the VLM's inference-time logits:
they are orders of magnitude flatter, with an Otsu threshold rank of
5,767 instead of 2. A decoder trained on raw proxy distributions and
deployed on VLM distributions would therefore face a sharp
train-inference mismatch. 
LC mitigates this mismatch by aligning four
distributional statistics (mean entropy, mean confidence,
95th-percentile entropy, and 95th-percentile confidence) through a
gradient-free scale, temperature, and label-smoothing transform
$(a,\alpha,\epsilon)$ applied to the proxy logits~\cite{guo2017calibration,muller2019label}.
Target statistics are
estimated once from the VLM over 100 reference prompts from the GenEval
benchmark~\cite{ghosh2023geneval}; the transform introduces no
learnable parameters and is fixed before decoder training begins. The
optimization procedure and full derivation are provided in
Appendix~\ref{sec:extended_lc}. Appendix~\ref{sec:lc_sensitivity}
further analyzes sensitivity to the number and domain of reference
prompts, showing that 100 GenEval prompts fall within a stable
calibration regime.
Although LC matches only global statistics, the calibrated proxies
agree with actual VLM logits position by position: their
distribution-weighted code vectors reach 0.86 cosine similarity with
those computed from the VLM, which drops to 0.04 when code identities
are permuted while all global statistics are kept
(Appendix~\ref{sec:proxy_validity}). The agreement therefore stems
from the shared codebook geometry of Section~\ref{sec:motivation},
which LC preserves because its transform cannot reorder a
distribution.

\subsection{Diffusion Decoder (DD)}
\label{sec:dd}
The conditioning representation $(v_i, u_i)$ is consumed by a
diffusion decoder~\cite{rombach2022latent,peebles2023scalable,labs2025flux} that 
operates in the latent space of a pre-trained VAE~\cite{kingma2013autoencoding}.
We finetune FLUX.1-Fill~\cite{labs2025flux},
a recent rectified-flow transformer~\cite{esser2024scaling,lipman2023flow,liu2023flow} for image inpainting, well-suited to localized,
token-aligned conditioning, by updating only its attention and
adapter layers. 
The distribution-weighted code vector $v_i$ is
spatially reshaped, projected through a linear layer and a $3 \times
3$ convolution, and added to the noisy latent tokens; uncertainty
features are bilinearly upsampled and concatenated channel-wise.
Training uses the $v$-prediction
objective~\cite{kingma2023understanding} on ImageNet-1K only, without
any text-to-image supervision. The decoder is conditioned only on
VLM-derived token representations and excludes direct text
conditioning by design, ensuring it follows the VLM's original token
belief rather than reinterpreting the scene with its own prior. Full
architecture and conditioning-injection details are in
Appendix~\ref{sec:extended_dd}.

\section{Experiments}
\label{sec:experiments}

\subsection{Setup}
We train the diffusion decoder on the ImageNet-1K training set
\cite{imagenet15russakovsky} and evaluate on ImageNet-1K validation
(in-domain) and MJHQ-30K \cite{li2024playground} (out-of-domain).
For reconstruction, we measure fidelity by rFID
\cite{heusel2017gans}, SSIM \cite{wang2004image}, and LPIPS
\cite{zhang2018unreasonable}. For text-to-image generation, we
measure fidelity by gFID on MJHQ-30K, and semantic preservation by
GenEval \cite{ghosh2023geneval}, DPG-Bench \cite{hu2024ella}, and
WISE \cite{niu2025wise}. Semantic-preservation metrics are reported
to verify that fidelity gains do not come at the cost of altering
the VLM's intent. We finetune FLUX.1-Fill \cite{labs2025flux} as the decoder, updating
only its attention and adapter layers, on 4$\times$A100 GPUs with
batch size 32 and 50K steps; inference uses 30 denoising steps. 
The default backbone is Janus-Pro 7B \cite{chen2025janus} at $384
\times 384$ image resolution; we additionally evaluate Janus-Pro 1B
and a second backbone, LlamaGen-XL ($256 \times 256$), in
Section~\ref{sec:regime}. Full hyperparameters and the
calibration procedure are in Appendix~\ref{sec:extended_imp}, and
the main results are stable across diffusion-noise seeds
(Appendix~\ref{sec:seeds}).

\begin{wraptable}{r}{0.5\textwidth}
    \vskip -0.5in
    \centering
    \small
    \setlength{\tabcolsep}{4pt}
    \renewcommand{\arraystretch}{0.95}
    \caption{\textbf{Image reconstruction on ImageNet-1K and MJHQ-30K} (Janus-Pro 7B). Reconstruction from VQ-VAE-encoded representations, comparing the native VQ-VAE decoder against DCDD. The decoder is trained on ImageNet-1K only, so the MJHQ-30K result tests cross-domain generalization. DCDD substantially reduces rFID across both datasets while consistently improving SSIM and LPIPS. Superior results are \hl{highlighted}.}
    
    \resizebox{0.5\textwidth}{!}{
    \begin{tabular}{lccc}
    \Xhline{4\arrayrulewidth}
        \rowcolor{table-yellow!60}\textit{\textbf{ImageNet-1K Val.}} & rFID↓ & SSIM↑ & LPIPS↓ \\
        \midrule
        VQ-VAE Decoder & 2.8511 & 0.5786 & 0.1541 \\
        \textbf{DCDD (ours)} & \hl{0.7535} & \hl{0.5957} & \hl{0.1374} \\
    \end{tabular}
    }
    \resizebox{0.5\textwidth}{!}{
    \begin{tabular}{lccc}
    \Xhline{3\arrayrulewidth}
        \rowcolor{table-yellow!60}\textit{\textbf{MJHQ-30K}} & rFID↓ & SSIM↑ & LPIPS↓ \\
        \midrule
        VQ-VAE Decoder & 4.4882 & 0.6564 & 0.1432 \\
        \textbf{DCDD (ours)} & \hl{1.8127} & \hl{0.6782} & \hl{0.1308} \\
    \Xhline{4\arrayrulewidth}
    \end{tabular}
    }
    \label{tab:recon}
    \vskip -0.3in
\end{wraptable}

\subsection{Reconstruction Fidelity}
\label{sec:recon}
Image reconstruction from VQ-VAE-encoded representations isolates
the decoder's contribution from token-prediction error and reflects
the fidelity ceiling of the decoder under correct token sequences.
Table~\ref{tab:recon} reports the result. DCDD reduces rFID by 74\%
on ImageNet-1K and 60\% on MJHQ-30K, with consistent improvements
in SSIM and LPIPS. The MJHQ-30K result, where training is performed
only on ImageNet-1K, indicates that the gain is not domain-locked.
This confirms that the VQ-VAE decoder, rather than the upstream
representation, is the dominant source of reconstruction-stage
fidelity loss. Qualitative samples are provided in Appendix
Figure~\ref{fig:recon_janus}.

\subsection{Text-to-Image: Fidelity and Semantic Preservation}
\label{sec:t2i}
We next apply DCDD to tokens generated by the VLM from text prompts.
Table~\ref{tab:t2i} reports per-category gFID on MJHQ-30K and
semantic alignment scores across three independent benchmarks
(GenEval, DPG-Bench, WISE). DCDD reduces gFID by 34\%
(14.29~$\to$~9.42) on MJHQ-30K, with consistent improvements across
nine of ten categories. By contrast, the semantic alignment metrics
show only small movement: GenEval overall +1.07, DPG-Bench overall
+0.70, and WISE Consistency +0.042. The larger sub-axis movements
within each benchmark (GenEval Single +1.25, Two +4.80; WISE Realism
+0.171, Aesthetic +0.250) are concentrated on axes that depend on
object-level visual detectability and rendering quality rather than
on alignment itself. Sharper boundaries, cleaner textures, and more
realistic appearance make objects more reliably recognized by
detection-based evaluators and rated as more visually plausible by
model-based evaluators, even when the semantics are unchanged. Qualitative samples are provided in Appendix
Figure~\ref{fig:t2i_janus}.

\begin{table}[t]
    \centering
\caption{\textbf{Text-to-image evaluation} (Janus-Pro 7B). DCDD is
designed to improve fidelity while preserving the VLM's semantic
intent. The fidelity benchmark (MJHQ-30K gFID) shows substantial
gains across nearly all categories. We report GenEval, DPG-Bench, and WISE as semantic consistency checks: because DCDD is a decoder-only fidelity module conditioned on the VLM's
own token belief, these scores are expected to remain stable rather than
increase substantially. Best per metric is \hl{highlighted}.}

    \resizebox{1.0\textwidth}{!}{
    \begin{tabular}{lccccccccccc}
    \Xhline{4\arrayrulewidth}
    \multicolumn{12}{c}{\textbf{\textit{Fidelity Metric ↓}}} \\
    \Xhline{2\arrayrulewidth}
     \rowcolor{table-yellow!60}\textit{\textbf{MJHQ-30K}} & \textbf{Overall} & Animals & Art & Fashion & Food & Indoor & Landscape & Logo & People & Plants & Vehicles \\
    \Xhline{2\arrayrulewidth}
        VQ-VAE Decoder & 14.2888 & 31.6866 & 35.6408 & 32.1410 & 33.1787 & 32.4016 & 30.5020 & 42.3565 & 37.6049 & \hl{33.1228} & 27.4424 \\
        \textbf{DCDD (ours)} & \hl{9.4160} & \hl{25.1060} & \hl{28.9806} & \hl{26.0182} & \hl{28.1580} & \hl{28.5172} & \hl{26.7104} & \hl{38.2032} & \hl{30.2217} & 33.6617 & \hl{25.0277} \\
    \Xhline{4\arrayrulewidth}
    \end{tabular}
    }
\resizebox{\linewidth}{!}{
\begin{tabular}{lccccccc}
\Xhline{4\arrayrulewidth}
\multicolumn{8}{c}{\textbf{\textit{Semantic Consistency Checks}}} \\
\Xhline{2\arrayrulewidth}
\rowcolor{table-yellow!60}
\textit{\textbf{GenEval}} & \textbf{Overall}
& Single & Two & Counting
& Colors  & Position & Color Attr. \\
\Xhline{2\arrayrulewidth}
VQ-VAE Decoder & 77.24 & 97.81 & 84.09 & 56.56 & 87.50 & \hl{75.75} & 61.75  \\
\textbf{DCDD (ours)} & \hl{78.31} & \hl{99.06} & \hl{88.89} & \hl{59.38} & \hl{88.30}
& 72.50 & 61.75 \\
\Xhline{2\arrayrulewidth}
\rowcolor{table-yellow!60}
\textit{\textbf{DPG-Bench}} & \textbf{Overall}
& Global & Relation & Entity
& Attribute & Others & \cellcolor{gray!100}--  \\
\Xhline{2\arrayrulewidth}
VQ-VAE Decoder & 86.94
& 82.37 & 93.38 & 89.76 & \hl{87.61} & 81.60 & \cellcolor{gray!100}--  \\
\textbf{DCDD (ours)} & \hl{87.64}
& \hl{83.59} & \hl{93.42} & \hl{90.05} & 87.15 & \hl{82.00} & \cellcolor{gray!100}--  \\
\Xhline{2\arrayrulewidth}
\rowcolor{table-yellow!60}
\textit{\textbf{WISE}}
& \textbf{Consistency} & Realism & Aesthetic Quality
& \cellcolor{gray!100}-- & \cellcolor{gray!100}-- & \cellcolor{gray!100}-- & \cellcolor{gray!100}-- \\
\Xhline{2\arrayrulewidth}
VQ-VAE Decoder
& 1.278 & 0.936 & 0.923 & \cellcolor{gray!100}-- & \cellcolor{gray!100}-- & \cellcolor{gray!100}-- & \cellcolor{gray!100}-- \\
\textbf{DCDD (ours)}
& \hl{1.320} & \hl{1.107} & \hl{1.173} & \cellcolor{gray!100}-- & \cellcolor{gray!100}-- & \cellcolor{gray!100}-- & \cellcolor{gray!100}--  \\
\Xhline{4\arrayrulewidth}
\end{tabular}
}
    \label{tab:t2i}
    \vskip -0.1
    in
\end{table}

\begin{wraptable}{r}{0.54\textwidth}
    \vskip -0.25in
    \centering
    \small
    \setlength{\tabcolsep}{4pt}
    \renewcommand{\arraystretch}{0.95}
\caption{\textbf{Image-to-image refinement vs. logit-level
conditioning on MJHQ-30K} (Janus-Pro 7B). gFID is computed against
MJHQ-30K reference images; CLIP-I and DINOv2 measure semantic
preservation of the refined output relative to the VQ-VAE-decoded
image (the VLM's intended output). \textit{I2I (generic)} uses a
fixed quality-improvement prompt (detailed in Figure~\ref{fig:qual}
caption); \textit{I2I + prompt} uses the
original MJHQ-30K generation prompt. Superior results are \hl{highlighted}.}
    \resizebox{0.5\textwidth}{!}{
    \begin{tabular}{lccc}
    \Xhline{4\arrayrulewidth}
        \rowcolor{table-yellow!60}\textit{\textbf{MJHQ-30K}} & gFID↓ & CLIP-I↑ & DINOv2↑ \\
        \midrule
        VQ-VAE & 14.2888 & \cellcolor{gray!100}-- & \cellcolor{gray!100}-- \\
        \textit{I2I (generic)} & 10.6678 & 0.8752 & 0.7663 \\
        \textit{I2I + prompt}    & 10.9767 & 0.8531 & 0.7586 \\
        \textbf{DCDD (ours)} & \hl{9.4160} & \hl{0.9088} & \hl{0.8692} \\
    \Xhline{4\arrayrulewidth}
    \end{tabular}
    }
    \label{tab:i2i}
    \vskip -0.3in
\end{wraptable}

\subsection{Logit-Level Conditioning vs. Image-to-Image Refinement}
\label{sec:llcvspsr}

\begin{figure}[t]
    \centering    
    \includegraphics[width=\textwidth]{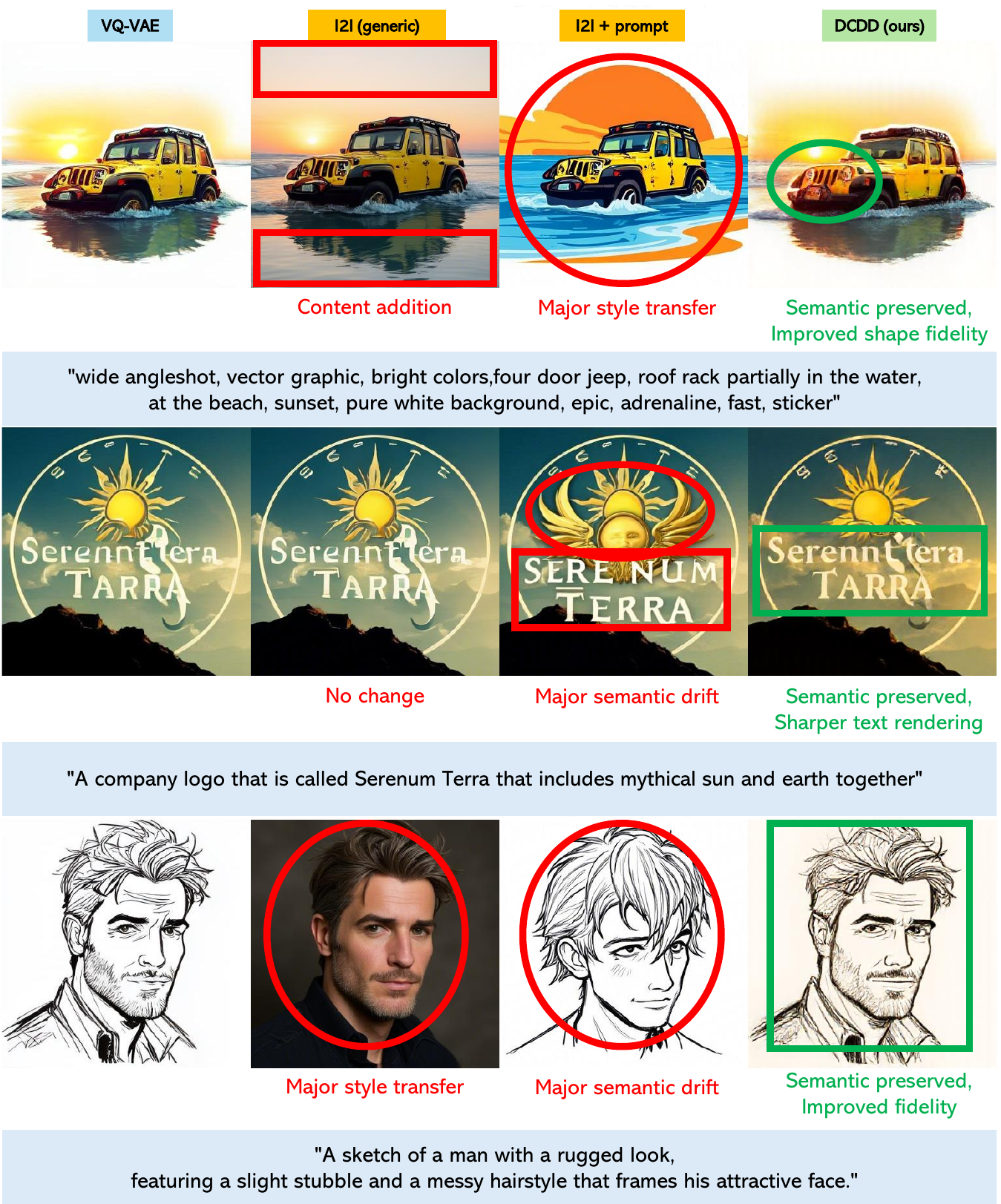}
    \caption{\textbf{Image-to-image refinement vs. our logit-conditioned
    decoding.} Qualitative comparison on Janus-Pro 7B generations from
    MJHQ-30K prompts. From left to right: native VQ-VAE decoding; FLUX.1-Kontext with a generic enhancement prompt (\textit{``enhance to
    high quality, preserve all original details and composition,
    sharpen fine details, same scene, same content, same colors, best
    quality''}); FLUX.1-Kontext conditioned on the original MJHQ-30K
    generation prompt; and DCDD. FLUX-based I2I refinement either
    under-modifies the image or alters the semantics substantially,
    while DCDD improves visual fidelity while preserving the VLM's
    semantic intent.}

   \label{fig:qual}
    \vskip -0.15in
\end{figure}

A natural alternative to DCDD is to leave the VQ-VAE decoder in
place and apply an off-the-shelf image-to-image (I2I) refiner to its
output. We compare against FLUX.1-Kontext~\cite{labs2025flux}, which
shares the same backbone family as our Diffusion Decoder, applied as
an I2I refiner in two variants: a generic enhancement pass with a
fixed quality-improvement prompt (\textit{I2I (generic)}) and a pass
that receives the original MJHQ-30K generation prompt
(\textit{I2I + prompt}). 
As shown in Figure~\ref{fig:qual}, this approach either under-modifies or alters scene content in ways that drift from the VLM's intent.

To quantify this trade-off, Table~\ref{tab:i2i} reports gFID
alongside two semantic-preservation metrics measured against the
VQ-VAE-decoded image (the VLM's intended output) on MJHQ-30K.
CLIP-I~\cite{radford2021learning} cosine similarity captures
high-level semantic agreement such as scene category and salient
objects; DINOv2~\cite{oquab2023dinov2} cosine similarity is more
sensitive to object identity and spatial layout, flagging cases
where the broad category is preserved but the specific subject has
drifted.

The I2I refiner also achieves a fidelity gain on gFID, but at the
cost of lower CLIP-I and DINOv2 similarity to the VQ-VAE output,
indicating semantic drift from the VLM's intent. Both I2I variants
exhibit similar drift regardless of prompt strategy, with DINOv2
nearly identical (0.7663 vs.\ 0.7586), suggesting the refiner's
structural rewrites are governed by its own pre-trained prior
rather than by the conditioning image. DCDD, in contrast, attains
the strongest fidelity improvement while preserving both similarity
metrics at higher values, demonstrating that logit-level
conditioning enhances visual quality without overwriting the VLM's
semantic intent.

\subsection{Component Contributions}
\label{sec:cctd}
A strong pretrained diffusion decoder may account for part of
the observed improvement. To separate this effect from that of
distributional conditioning, we train a baseline diffusion decoder
conditioned only on the code vectors of sampled tokens.
This baseline uses the same FLUX.1-Fill initialization, trainable
layers, training data, optimizer, batch size, 50K training steps,
and inference schedule as DCDD, differing only in its conditioning.
As shown in Table~\ref{tab:argmax}, the diffusion decoder alone
reduces gFID from 14.29 to 11.08, while full-distribution
conditioning further reduces it to 9.42.
This corresponds to a 15\% improvement over the argmax-conditioned
baseline, with a similar benefit in reconstruction
(36\% lower rFID on ImageNet-1K).
Since the two decoders share the same architecture and training
budget, the additional improvement reflects the contribution
of LCDM and LC rather than increased decoder capacity.
Most of the GenEval improvement is already achieved with argmax
conditioning, consistent with the detectability effect discussed
in Section~\ref{sec:t2i}.
We also evaluate a lightweight learned adapter that feeds soft
codes into the frozen native decoder (Appendix~\ref{sec:adapter}).
Although it improves reconstruction, it does not improve generation,
suggesting that directly adapting the frozen native decoder is
insufficient to fully exploit token distributions.

\textbf{How much of the distribution is needed?}
To examine the contribution of lower-ranked tokens, we retain
only the top-$k$ probabilities at each position and recompute
$(v_i, u_i)$, keeping the decoder and generated token sequences fixed
(Table~\ref{tab:topk}).
Including the second-ranked token recovers 76\% of the improvement
between top-1 and full-distribution conditioning.
Tokens ranked 3--8 account for another 18\%, while the remaining
tail contributes the rest.
This is consistent with the Otsu analysis in
Section~\ref{sec:motivation}, which identifies rank 2 as the
boundary between informative tokens and the noise floor.
Since the decoder is trained on full distributions, this analysis
serves as a diagnostic rather than an alternative inference
configuration.

\begin{figure}[h]
  \centering
   \includegraphics[width=1.0\linewidth]{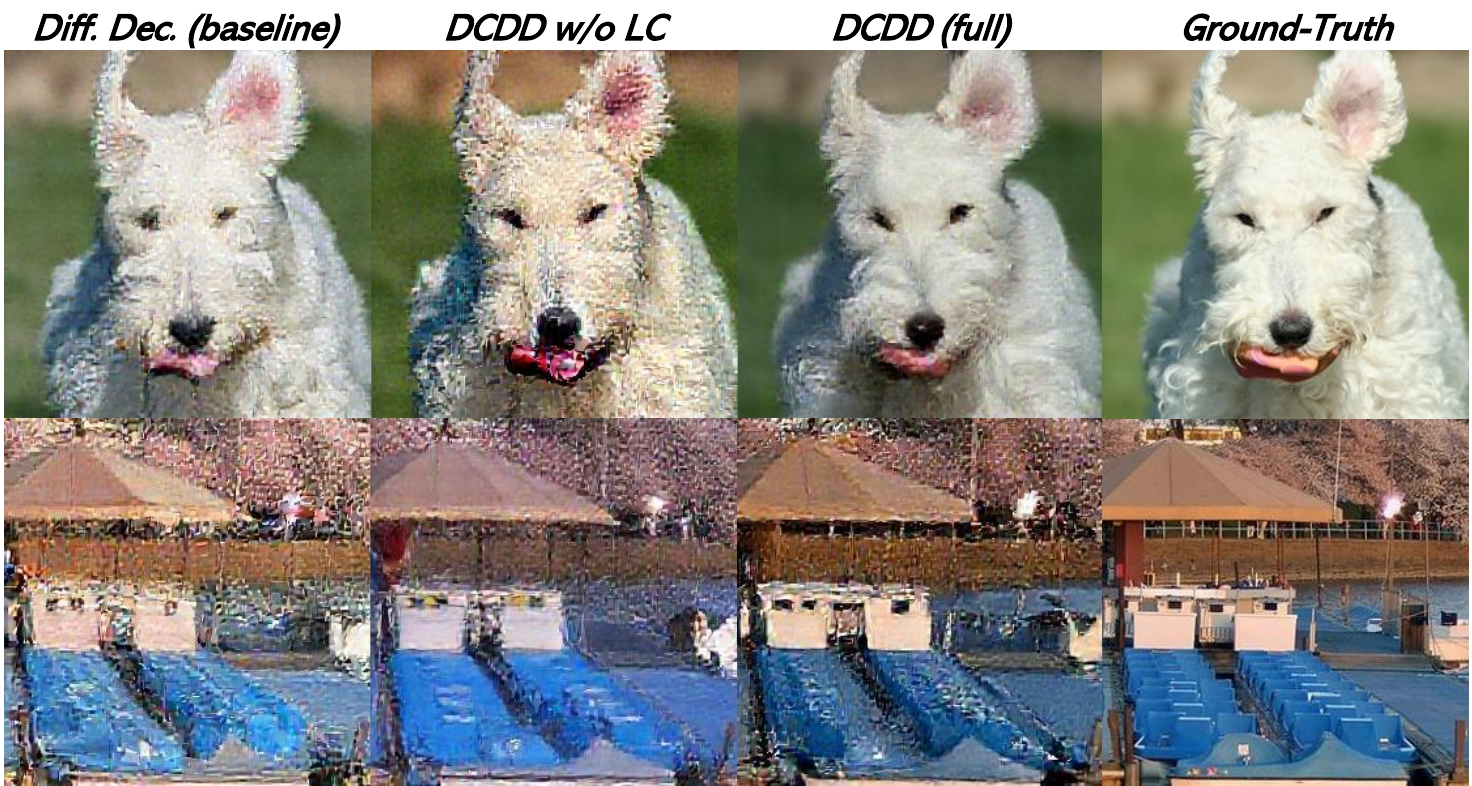}
\caption{\textbf{Component ablation on ImageNet-1K reconstruction at
10K training steps.} \textit{(a)} Discrete conditioning
(\textbf{\textit{Diff.\ Dec.\ (baseline)}}) produces low-frequency
noise and broken textures. \textit{(b)} Adding continuous
distribution-weighted conditioning (\textbf{\textit{DCDD w/o LC}})
recovers structure but inflates contrast and introduces residual
color noise. \textit{(c)} Adding calibration
(\textbf{\textit{DCDD (full)}}) stabilizes color tones and
suppresses noise. Each transition addresses a specific failure mode
rather than incrementally improving along a single axis.
}
\label{fig:components}
    \vskip -0.2in
\end{figure}

Finally, stepwise ablations (Appendix~\ref{sec:stepwise})
show that LCDM contributes most to the rFID reduction, while LC
provides the largest SSIM gain.
Figure~\ref{fig:components} illustrates the distinct failure
modes addressed by each component.

\begin{table}[t]
\centering
\begin{minipage}[t]{0.56\textwidth}
\centering
\small
\captionof{table}{\textbf{Isolating distributional conditioning}
(Janus-Pro 7B). Both diffusion decoders share initialization,
trainable layers, data, and the 50K-step schedule; only the
conditioning differs. Superior results are \hl{highlighted}.}
\label{tab:argmax}
\resizebox{\linewidth}{!}{
\begin{tabular}{lcccc}
\Xhline{4\arrayrulewidth}
\rowcolor{table-yellow!60}
& \multicolumn{2}{c}{\textbf{rFID}$\downarrow$} & \textbf{gFID}$\downarrow$ & \textbf{GenEval}$\uparrow$ \\
\rowcolor{table-yellow!60}
\textit{\textbf{Conditioning}} & IN-1K & MJHQ & MJHQ & Overall \\
\midrule
VQ-VAE Decoder & 2.8511 & 4.4882 & 14.2888 & 77.24 \\
Diff.\ Dec.\ (argmax codes) & 1.1842 & 2.4325 & 11.0837 & 78.08 \\
\textbf{DCDD (full distribution)} & \hl{0.7535} & \hl{1.8127} & \hl{9.4160} & \hl{78.31} \\
\Xhline{4\arrayrulewidth}
\end{tabular}
}
\end{minipage}
\hfill
\begin{minipage}[t]{0.425\textwidth}
\centering
\small
\captionof{table}{\textbf{Top-$k$ truncation at inference}
(Janus-Pro 7B, fixed decoder and tokens). gFID on the MJHQ-30K
Animals category.}
\label{tab:topk}
\resizebox{\linewidth}{!}{
\begin{tabular}{lcc}
\Xhline{4\arrayrulewidth}
\rowcolor{table-yellow!60}
\textit{\textbf{Condition}} & gFID$\downarrow$ & GenEval$\uparrow$ \\
\midrule
VQ-VAE Decoder & 31.6866 & 77.24 \\
Top-1 only & 33.2015 & 76.07 \\
Top-2 & 27.0649 & 77.83 \\
Top-8 & 25.5681 & 77.90 \\
\textbf{Full (DCDD)} & \hl{25.1060} & \hl{78.31} \\
\Xhline{4\arrayrulewidth}
\end{tabular}
}
\end{minipage}
\vskip -0.27in
\end{table}

\begin{wraptable}{r}{0.5\textwidth}
    \vskip -0.45in
    \centering
    \small
    \setlength{\tabcolsep}{4pt}
    \renewcommand{\arraystretch}{0.95}
    \caption{\textbf{End-to-end T2I cost on MJHQ-30K} (Janus-Pro 7B,
    batch size 1, A100 80GB). DCDD adds moderate inference latency
    for a substantial fidelity gain. The native VQ-VAE path remains
    available when efficiency is preferred.}
    \resizebox{0.5\textwidth}{!}{
    \begin{tabular}{lcc}
    \Xhline{4\arrayrulewidth}
        \rowcolor{table-yellow!60}
        \textit{\textbf{Configuration}}
        & gFID $\downarrow$ & Latency (s) $\downarrow$ \\
    \Xhline{2\arrayrulewidth}
        VLM + VQ-VAE
            & 14.28 & 11.86 \\
        VLM + DCDD (ours)
            & 9.41 & 15.56 \\
    \midrule
        \textit{Trade-off}
            & {\color{green!60!black}$-$34.12\%}
            & {\color{red!70!black}$+$31.19\%} \\
    \Xhline{4\arrayrulewidth}
    \end{tabular}
    }
    \label{tab:computation}
    \vskip -0.1in
\end{wraptable}

\subsection{Computational Cost}
\label{sec:cost}
With 30 denoising steps, DCDD adds 31\% end-to-end latency for a
34\% gFID reduction on MJHQ-30K (Table~\ref{tab:computation}). The
end-to-end cost is dominated by the VLM rather than
the decoder, and the native VQ-VAE path remains available when
efficiency is preferred. On the training side, the decoder is
trained on ImageNet-1K alone for 50K steps on 4$\times$A100; the
VLM and VAE are not modified, in contrast to tokenizer-replacement
and joint-optimization alternatives that require foundation-scale
T2I retraining. 
Even with 10 steps, DCDD outperforms the native decoder in reconstruction at only 10.7\% additional latency (Appendix~\ref{sec:latency}).

\subsection{Regime of Applicability}
\label{sec:regime}
A post-hoc decoder can only exploit information available in its
input. We therefore examine two additional models
(Table~\ref{tab:regime}). Janus-Pro 1B~\cite{chen2025janus}
shares the tokenizer with Janus-Pro 7B, allowing us to reuse
the 7B-trained DCDD checkpoint without retraining.
We align the 1B logits to the 7B statistics using the same
gradient-free procedure as LC (Appendix~\ref{sec:janus1b}).
With similar logit characteristics (Otsu rank 2, top-1 0.25),
DCDD reduces gFID by 25.7\% on MJHQ-30K Animals, compared
with 20.8\% for Janus-Pro 7B, while GenEval decreases by
only 1.3 points. In contrast, LlamaGen-XL has a more concentrated distribution (Otsu rank 1) and produces weaker token sequences. DCDD provides no
text-to-image improvement, although reconstruction gains
remain (Appendix~\ref{sec:appendix_vlm}).
These results suggest that DCDD is most effective when
upstream token sequences are semantically coherent and
the logits retain meaningful probability mass beyond rank 1.

\begin{table}[t]
\centering
\small
\caption{\textbf{Regime of applicability.} VLM logit statistics and the corresponding effect of DCDD on generation performance (VQ-VAE $\to$ DCDD). gFID is evaluated on MJHQ-30K Animals for all models. $^\dagger$Janus-Pro 1B uses the 7B-trained DCDD checkpoint without retraining.
}
\label{tab:regime}
\resizebox{0.99\linewidth}{!}{
\begin{tabular}{lccccc}
\Xhline{4\arrayrulewidth}
\rowcolor{table-yellow!60}
\textit{\textbf{Model}} & Otsu Rank & Top-1 & Norm.\ Entropy & gFID$\downarrow$ (Animals) & GenEval$\uparrow$ \\
\midrule
Janus-Pro 7B & 2 & 0.2216 & 0.4153 & 31.69 $\to$ \textbf{25.11} & 77.24 $\to$ 78.31 \\
Janus-Pro 1B $^\dagger$ & 2 & 0.2457 & 0.4033 & 35.14 $\to$ \textbf{26.09} & 63.46 $\to$ 62.17 \\
LlamaGen-XL & 1 & 0.3794 & 0.3278 & 74.20 $\to$ 74.66 & 30.02 $\to$ 28.19 \\
\Xhline{4\arrayrulewidth}
\end{tabular}
}
\vskip -0.2in
\end{table}

\subsection{Additional Analyses}
\label{sec:additional_analyses}

Several robustness and diagnostic analyses are deferred to the
appendix. Appendix~\ref{sec:proxy_validity} verifies that calibrated
proxies and VLM logits induce compatible decoder conditioning, and
Appendix~\ref{sec:stability} analyzes the stability of that
conditioning; Appendix~\ref{sec:appendix_denoisingsteps} and
Appendix~\ref{sec:latency} examine the number of denoising steps,
latency, and memory; Appendix~\ref{sec:lc_sensitivity} verifies that
Logit Calibration is stable across the size and content of its
reference prompt set, ruling out benchmark-specific bias;
Appendix~\ref{sec:appendix_preliminary} confirms that LC reshapes
the encoder's probability decay to closely track the VLM's, as
motivated in Section~\ref{sec:motivation}; and
Appendix~\ref{sec:seeds} reports variation across diffusion-noise
seeds.

\section{Conclusion}
\label{sec:conclusion}
We presented DCDD, a post-hoc decoder framework built on a simple
observation about pre-trained discrete-token VLMs: before sampling, the
VLM's logit distribution carries rank-2 or higher codebook information
that standard decoding discards after selecting a single token index.
DCDD combines Logit-to-Code Distributional Mapping, gradient-free Logit
Calibration, and a Diffusion Decoder to recover this discarded signal
as a high-fidelity image, serving as an optional alternative to the
native VQ-VAE decoder while leaving the VLM frozen. Trained on
ImageNet-1K alone, DCDD substantially improves reconstruction and
generation fidelity while keeping semantic consistency stable. More
broadly, our results show that useful distributional information can be
recovered post-hoc from a frozen discrete-token model's logits.

\clearpage

\begin{ack}
This research was supported by the MSIT(Ministry of Science, ICT), Korea, under the Global Research Support Program in the Digital Field program(RS-2024-00436680) supervised by the IITP(Institute for Information \& Communications Technology Planning \& Evaluation). This project is supported by Microsoft Research Asia.
The authors declare no competing financial interests beyond those disclosed above.
\end{ack}

{
\small
\bibliographystyle{plain}
\bibliography{bibliography}
}

\clearpage
\appendix

\section{Extended Method Details}
\label{sec:extended_method}

\begin{figure}[h]
  \centering
   \includegraphics[width=1.0\linewidth]{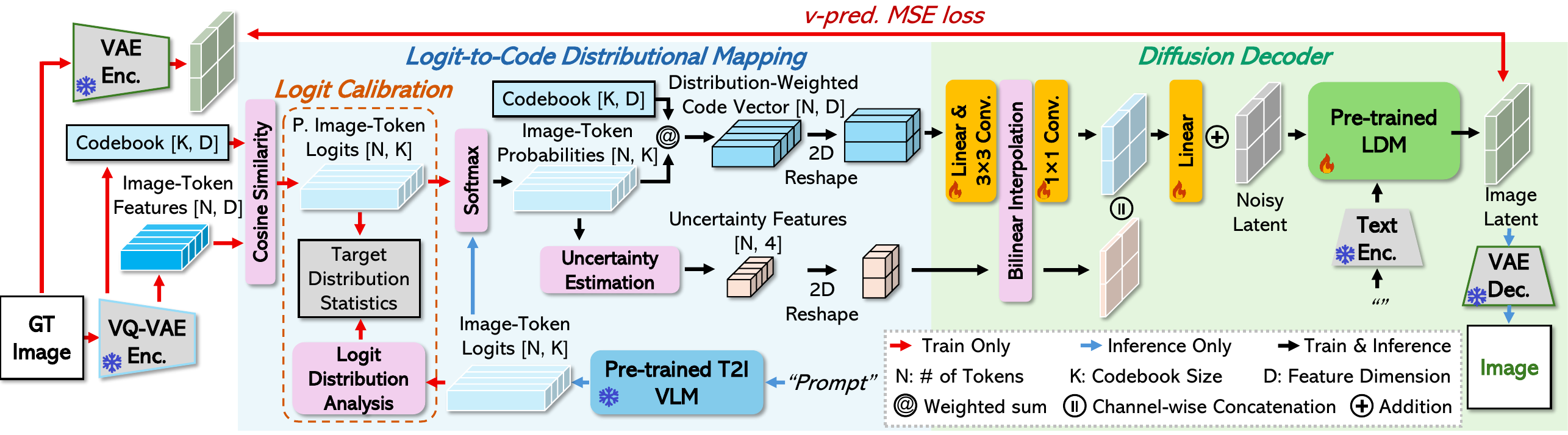}
\caption{\textbf{Detailed view of the DCDD framework.} An expanded
version of Figure~\ref{fig:overview} showing per-stage tensor shapes
and explicit train/inference data flow. \textit{Inference:} the
frozen VLM produces image-token logits, which LCDM converts into
continuous distribution-weighted code vectors and uncertainty
features that condition the Diffusion Decoder.
\textit{Training:} ground-truth images pass through the frozen
VQ-VAE encoder; cosine similarity to the codebook produces proxy
image-token logits, which gradient-free Logit Calibration aligns
to the VLM's logit statistics before feeding the same LCDM stage.
}

   \label{fig:overview_detail}
\vskip -0.15in
\end{figure}

This section provides the full formulation of DCDD's three components,
expanding on the main-paper description in Section~\ref{sec:method}.
Figure~\ref{fig:overview_detail} provides a detailed view of the
framework, showing the per-stage tensor shapes and the train/inference
data flow. The remainder of the section presents Logit-to-Code
Distributional Mapping (Appendix~\ref{sec:extended_lcdm}), Logit
Calibration (Appendix~\ref{sec:extended_lc}), and the Diffusion Decoder
(Appendix~\ref{sec:extended_dd}) in turn.

\subsection{Logit-to-Code Distributional Mapping}
\label{sec:extended_lcdm}
This stage transforms the image-token logits generated from a
pre-trained autoregressive VLM into continuous conditioning
representations that enrich the diffusion decoder with dense,
fine-grained visual information for high-fidelity image synthesis.

Given either the \textbf{\emph{Image-Token Logits}} from the VLM
during inference or the \textbf{\emph{Proxy Image-Token Logits}}
derived from the VQ-VAE encoder during training, this mapping
converts these logits into two key representations: the
\textbf{\emph{Distribution-Weighted Code Vector}} capturing the
expected semantic embedding per token, and the complementary
\textbf{\emph{Uncertainty Features}} quantifying confidence and
ambiguity in token prediction.

\textbf{Inference with VLM:}

During inference, the autoregressive VLM produces a sequence of
\textbf{\emph{Image-Token Logits}}
\begin{equation}
\mathbf{L} = [\mathbf{l}_1, \ldots, \mathbf{l}_N] \in
\mathbb{R}^{N \times K},
\end{equation}
where $N$ denotes the number of image tokens and $K$ the size of
the visual codebook. Each $\mathbf{l}_i = [l_{i,1}, \ldots, l_{i,K}]$
contains scores over the $K$ \textbf{Code Vectors}
$\mathcal{E} = \{\mathbf{e}_1, \ldots, \mathbf{e}_K\}$ with
$\mathbf{e}_k \in \mathbb{R}^D$, produced by the VLM's output
projection from its hidden state.
In practice, the logits may be obtained after applying classifier-free
guidance and/or temperature scaling; for notational simplicity, we
omit these operators and write the resulting distribution in a
standard softmax form.

The logits are transformed into normalized
\textbf{\emph{Image-Token Probabilities}} via a standard softmax:
\begin{equation}
p_{i,k} = \frac{\exp(l_{i,k})}{\sum_{j=1}^{K} \exp(l_{i,j})},
\quad
\mathbf{P} = [p_{i,k}] \in \mathbb{R}^{N \times K}.
\label{eq:itp}
\end{equation}

Each row of $\mathbf{P}$ represents the probability distribution
over the $K$ codebook entries for a specific spatial token position.
To convert these distributions into continuous embeddings, each
probability vector $\mathbf{p}_i \in \mathbb{R}^{K}$ is used to
compute a weighted combination of the corresponding \textbf{Code
Vectors} in the codebook
$\mathbf{E} = [\mathbf{e}_1; \ldots; \mathbf{e}_K] \in
\mathbb{R}^{K \times D}$:
\begin{equation}
\mathbf{v}_i = \sum_{k=1}^{K} p_{i,k}\,\mathbf{e}_k,
\quad
\mathbf{v}_i \in \mathbb{R}^{D}.
\label{eq:v}
\end{equation}
Stacking all token embeddings yields the matrix form:
\begin{equation}
\mathbf{V} = \mathbf{P}\,\mathbf{E} \in \mathbb{R}^{N \times D}.
\label{eq:vs}
\end{equation}

This operation performs a probability-weighted aggregation of the
codebook features, projecting each token's probability distribution
$\mathbf{p}_i$ into the continuous latent space of the VQ-VAE
codebook. The resulting \textbf{\emph{Distribution-Weighted Code
Vectors}} $\mathbf{V}$ preserve the distributional information
contained in the VLM's predictions.

However, this weighted aggregation inherently compresses each
token's probability distribution into a single expectation,
discarding information about its sharpness and variability. To
compensate for this loss and explicitly capture the model's
confidence and ambiguity, we derive four scalar statistics from
$\mathbf{P}$ for each spatial token $i$:
\begin{align}
u^{(1)}_i &= \max_k p_{i,k} && \text{(confidence)} , \nonumber \\
u^{(2)}_i &= p_{i,(1)} - p_{i,(2)} && \text{(margin)} , \nonumber \\
u^{(3)}_i &= \textstyle\sum_{k \in \text{Top-}K_u} p_{i,k} &&
\text{(top-$K_u$ mass)} , \nonumber \\
u^{(4)}_i &= -\frac{\sum_{k \neq k^*} p_{i,k} \log p_{i,k}}
{\log (K-1)} && \text{(tail entropy)} ,
\label{eq:uncertainty}
\end{align}
where $p_{i,(1)}$ and $p_{i,(2)}$ denote the top-1 and top-2
probabilities, and $k^*=\arg\max_k p_{i,k}$. For the top-$K_u$ mass,
we employ an entropy-adaptive $K_u$ constrained between 8 and 64
tokens, ensuring that the statistic reflects an appropriate portion
of the probability mass across varying distribution sharpness.
These measures are concatenated into \textbf{\emph{Uncertainty
Features}} $\mathbf{U} \in \mathbb{R}^{N \times 4}$, which
complement the continuous code representations $\mathbf{V}$ by
summarizing the dispersion characteristics of each token's
probability distribution. Together, $(\mathbf{V}, \mathbf{U})$
provide the diffusion decoder with both semantic and
confidence-aware information.

\textbf{Training with VQ-VAE Encoder:}

During training, we generate \textbf{\emph{Proxy Image-Token
Logits}} using the \textbf{\emph{VQ-VAE encoder}} from the target
VLM's tokenizer. Given a ground-truth image $x \in \mathbb{R}^{3
\times H \times W}$, the VQ-VAE encoder downsamples the image
through a series of convolutional and residual blocks, producing
spatial latent features immediately before vector quantization.
These latent features, referred to as \textbf{\emph{Image-Token
Features}} $\mathbf{F}$, capture compact yet semantically rich
representations of local regions, encoding their color, texture,
and structural patterns. Thus, we obtain
\begin{equation}
\mathbf{F} = [\mathbf{f}_1, \ldots, \mathbf{f}_N] \in
\mathbb{R}^{N \times D},
\end{equation}
from the encoder, where $N$ is the number of spatial tokens and
$D$ is the feature dimension.

The original VQ-VAE pipeline selects each $\mathbf{f}_i$'s nearest
codebook entry by L2 distance. To produce a continuous distribution
over the codebook rather than a single index, we instead compute
cosine similarity between $\mathbf{f}_i$ and each entry in the
learned \textbf{\emph{Codebook}}
$\mathcal{E} = \{\mathbf{e}_1, \ldots, \mathbf{e}_K\}$ with
$\mathbf{e}_k \in \mathbb{R}^D$, yielding the
\textbf{\emph{Proxy Image-Token Logits}}:
\begin{equation}
s_{i,k} = \cos(\mathbf{f}_i, \mathbf{e}_k),
\quad
\mathbf{S} = [s_{i,k}] \in \mathbb{R}^{N \times K}.
\end{equation}

We then apply the same LCDM transform as in inference
(Eqs.~\eqref{eq:itp}--\eqref{eq:uncertainty}) to obtain the
\textbf{\emph{Distribution-Weighted Code Vectors}} and
\textbf{\emph{Uncertainty Features}} for diffusion conditioning:
\[
\mathbf{V}^{\text{enc}} = \mathbf{P}^{\text{enc}} \mathbf{E}
\in \mathbb{R}^{N \times D},
\qquad
\mathbf{U}^{\text{enc}} \in \mathbb{R}^{N \times 4}.
\]

For diffusion training, it is essential to establish explicit
correspondence between each ground-truth image and its
conditioning representation. The above process enables this by
constructing \textbf{\emph{Proxy Image-Token Logits}} that closely
align with the inference-time outputs of the VLM. Although these
proxy logits share the same codebook structure as the VLM's
generated logits, their probability distributions may diverge,
leading to a potential train-inference gap. To mitigate this
mismatch, we apply \textbf{Logit Calibration}, as described in the
following section.

\subsection{Logit Calibration}
\label{sec:extended_lc}
Although the \textbf{\emph{Proxy Image-Token Logits}} produced from
the VQ-VAE encoder live in the same codebook space as the VLM's
inference-time logits, their statistical behavior differs
significantly from that of the pre-trained VLM. In particular, the
encoder's cosine-similarity-based logits tend to exhibit
substantially higher entropy and weaker confidence than the VLM's
image-token logits, which encode a selective belief over a small
set of semantically relevant codebook entries (Section~\ref{sec:motivation}).
To bridge this discrepancy, we introduce a \textbf{Logit Calibration}
procedure that aligns the encoder's token-probability distribution
with the VLM's intrinsic confidence and entropy characteristics.

\paragraph{Target Distribution Statistics.}
We first estimate the \textbf{\emph{Target Distribution Statistics}}
of the VLM by sampling the pre-trained text-to-image model over a
set of random prompts. Specifically, we draw $N=100$ prompts from
the GenEval benchmark and generate $T=576$ image tokens per prompt
using classifier-free guidance~\cite{ho2022classifierfree} with weight $w_{\text{cfg}}=5.0$ and
temperature $\tau_{\text{VLM}}=1.0$. For each decoding step, we
obtain the guided image-token distribution $p^{\text{VLM}}(\cdot)$
and compute its token-level confidence $\max_k p_k$ and normalized
entropy
\begin{equation}
H_{\text{norm}}(p)
=
- \frac{1}{\log K}\sum_{k=1}^{K} p_k \log p_k,
\end{equation}
where $K=16{,}384$ is the size of the image-token vocabulary.
Aggregating these quantities across all $N \times T$ tokens yields
the VLM's target statistics
\begin{equation}
\mathcal{T}_{\text{VLM}}
=
\big(
\mu_H,\;
\mu_C,\;
P_{95}(C),\;
P_{95}(H)
\big),
\label{eq:target_stats}
\end{equation}
where $\mu_H$ and $P_{95}(H)$ denote the mean and 95th-percentile
entropy, and $\mu_C$ and $P_{95}(C)$ denote the corresponding
confidence statistics.

\paragraph{Encoder Logits and Calibration Transform.}
On the encoder side, we evaluate the VQ-VAE encoder on a set of
$M=100$ training images. For each latent image-token feature
$\mathbf{f}_i \in \mathbb{R}^D$, we compute the cosine-similarity
logits
\begin{equation}
s_{i,k}
=
\cos(\mathbf{f}_i, \mathbf{e}_k),
\end{equation}
where $\mathbf{e}_k \in \mathbb{R}^D$ denotes the $k$-th codebook
vector. These logits are then transformed via affine scaling and a
temperature-controlled softmax:
\begin{equation}
\tilde{s}_{i,k}
=
a\, s_{i,k},
\qquad
p^{\text{enc}}_{i,k}
=
\frac{\exp(\tilde{s}_{i,k}/\alpha)}
{\sum_{j=1}^{K}\exp(\tilde{s}_{i,j}/\alpha)},
\label{eq:calib_softmax}
\end{equation}
where $a$ denote the global logit scale, and
$\alpha$ controls the sharpness of the resulting distribution. After
softmax, we apply label smoothing
\begin{equation}
\hat{p}_{i,k}
=
(1-\epsilon)\,p^{\text{enc}}_{i,k}
+
\frac{\epsilon}{K},
\end{equation}
which mitigates overconfident predictions and stabilizes the tail
behavior of the distribution.

\paragraph{Calibration Objective and Optimization.}
From the calibrated encoder distributions $\hat{p}_{i,k}$, we compute
the corresponding entropy and confidence statistics
$\mu_H^{\text{enc}}, \mu_C^{\text{enc}}, P_{95}(C)^{\text{enc}},
P_{95}(H)^{\text{enc}}$ and match them to the VLM's target statistics
in Eq.~\eqref{eq:target_stats}. To quantify the mismatch, we define
the weighted objective
\begin{equation}
\begin{aligned}
\mathcal{L}(\theta)
&=
w_1(\mu_H^{\text{enc}}-\mu_H)^2
+
w_2(\mu_C^{\text{enc}}-\mu_C)^2
\\
&\quad+
w_3\!\left(P_{95}(C)^{\text{enc}}-P_{95}(C)\right)^2
\\
&\quad+
w_4\!\left(P_{95}(H)^{\text{enc}}-P_{95}(H)\right)^2,
\end{aligned}
\end{equation}
where $\theta=(a,\alpha,\epsilon)$ and
$(w_1,w_2,w_3,w_4)=(1.0,\,0.25,\,0.25,\,0.35)$. We prioritize
matching the mean entropy, which provides the most reliable
characterization of distributional sharpness, while confidence-based
terms serve as secondary constraints.

The calibration parameters are determined via a coordinated,
gradient-free search. We first tune the global logit scale $a$ over
$[1,\,60]$ using bisection until the encoder's mean entropy
satisfies $|\mu_H^{\text{enc}}-\mu_H|\le 0.02$, fixing
$(\alpha,\epsilon)=(\,1.0,\,0)$. Next, we sweep
$\alpha \in [0.5,\,2.0]$ to refine the distribution's softness,
followed by a search over $\epsilon \in [0.0,\,0.05]$ to align the
95th-percentile confidence. 

For Janus-Pro 7B, the final calibrated configuration is
$a=29.0381$, $\alpha=1.0$, and $\epsilon=0.01$. The calibration
transform is applied to the encoder-derived proxy logits during
training so that the diffusion decoder sees conditioning distributions
that match the VLM's inference-time statistics.
At inference, we use the VLM's image-token logits directly under the
same classifier-free guidance and sampling settings used to estimate
the target statistics ($w_{\text{cfg}}=5.0$,
$\tau_{\text{VLM}}=1.0$). This ensures that the decoder is trained on
calibrated proxy distributions that match the statistical regime of the
VLM logits it receives at deployment.

\subsection{Diffusion Decoder}
\label{sec:extended_dd}

The diffusion decoder is finetuned from FLUX.1-Fill, a rectified-flow
transformer for image inpainting~\cite{labs2025flux} built on the
flow-matching and rectified-flow generative modeling paradigm
\cite{lipman2023flow,liu2023flow}.
We update only its attention and adapter layers; all
other parameters are frozen. The decoder is conditioned only on
VLM-derived token representations and excludes direct text
conditioning by design, ensuring it follows the VLM's original token
belief rather than reinterpreting the scene with its own prior.

The decoder operates in the latent space of a pre-trained VAE.
Given an image $x \in \mathbb{R}^{3 \times H \times W}$, the VAE
encoder produces a latent representation
\begin{equation}
\mathbf{z}_{\text{img}} = \mathrm{VAE}_{\text{enc}}(x),
\quad
\mathbf{z}_{\text{img}} \in \mathbb{R}^{H' \times W' \times D_z},
\end{equation}
where $H'$ and $W'$ are the spatial dimensions of the latent feature
map and $D_z$ is the latent channel dimension. Noise is added
according to a timestep-dependent variance schedule:
\begin{equation}
\tilde{\mathbf{z}}_{\text{img},t}
= (1 - \sigma_t)\,\mathbf{z}_{\text{img}} + \sigma_t\,\boldsymbol{\epsilon},
\quad
\boldsymbol{\epsilon} \sim \mathcal{N}(\mathbf{0}, \mathbf{I}),
\end{equation}
where $\sigma_t$ is the noise variance coefficient at timestep $t$.

For conditioning, the \textbf{\emph{Distribution-Weighted Code
Vectors}} $\mathbf{V}^{\text{enc}} \in \mathbb{R}^{N \times D}$ and
the \textbf{\emph{Uncertainty Features}}
$\mathbf{U}^{\text{enc}} \in \mathbb{R}^{N \times 4}$ are spatially
reshaped and interpolated to match the latent resolution. The code
vectors are linearly projected and then refined through a lightweight
convolutional layer:
\begin{equation}
\mathbf{H}_{\text{code}}
= \phi_{\text{conv}}\big(\mathrm{Linear}(\mathbf{V}^{\text{enc}})\big),
\quad
\mathbf{H}_{\text{code}} \in \mathbb{R}^{N \times D_h},
\end{equation}
where $\mathrm{Linear}(\cdot)$ maps each code vector from
$\mathbb{R}^{D}$ to a hidden dimension $\mathbb{R}^{D_h}$ aligned
with the diffusion transformer input, and $\phi_{\text{conv}}(\cdot)$
is a $3{\times}3$ convolution with normalization and activation that
enhances local spatial coherence among neighboring tokens. The
uncertainty features $\mathbf{U}^{\text{enc}}$ are bilinearly
upsampled to the same spatial resolution and concatenated
channel-wise with the refined code features, forming the conditioning
tensor:
\begin{equation}
\mathbf{C} = [\mathbf{H}_{\text{code}}, \mathbf{U}^{\text{enc}}],
\quad
\mathbf{C} \in \mathbb{R}^{N' \times (D_h + 4)},
\end{equation}
where $N'$ is the number of spatial tokens after interpolation.

The noisy latent $\tilde{\mathbf{z}}_{\text{img},t}$ is divided into
patch tokens and projected into the transformer feature space:
\begin{align}
\mathbf{H}_{\text{img}} &=
\mathrm{Proj}_{\text{img}}(\mathrm{Pack}(\tilde{\mathbf{z}}_{\text{img},t})),
\quad
\mathbf{H}_{\text{img}} \in \mathbb{R}^{N' \times D_\theta}, \\
\mathbf{H}_{\text{cond}} &=
\mathrm{Proj}_{\text{cond}}(\mathbf{C}),
\quad
\mathbf{H}_{\text{cond}} \in \mathbb{R}^{N' \times D_\theta},
\end{align}
where $\mathrm{Pack}(\cdot)$ flattens the latent map into a sequence
of $N'$ spatial tokens, and $\mathrm{Proj}_{\text{img}}(\cdot)$ and
$\mathrm{Proj}_{\text{cond}}(\cdot)$ are learned linear projections
mapping the respective features into the transformer's internal
feature dimension $D_\theta$. The two representations are fused by
element-wise addition:
\begin{equation}
\mathbf{H}_{\text{input}} = \mathbf{H}_{\text{img}} + \mathbf{H}_{\text{cond}},
\quad
\mathbf{H}_{\text{input}} \in \mathbb{R}^{N' \times D_\theta}.
\end{equation}

The fused tokens are processed by the diffusion transformer:
\begin{equation}
\mathbf{H}_{\text{out}}^{t} = \mathcal{D}_{\theta}(\mathbf{H}_{\text{input}}, t),
\quad
\mathbf{H}_{\text{out}}^{t} \in \mathbb{R}^{N' \times D_\theta},
\end{equation}
where $\mathcal{D}_{\theta}$ is the diffusion transformer
parameterized by $\theta$. The output features are projected back to
the latent channel dimension and reshaped into the spatial latent
form to predict the residual velocity:
\begin{equation}
\hat{\mathbf{v}}^{t}
= \mathrm{Unpack}\!\left(\mathrm{Proj}_{\text{out}}(\mathbf{H}_{\text{out}}^{t})\right),
\quad
\hat{\mathbf{v}}^{t} \in \mathbb{R}^{H' \times W' \times D_z},
\end{equation}
where $\mathrm{Proj}_{\text{out}}(\cdot)$ projects to the latent
dimension $D_z$ and $\mathrm{Unpack}(\cdot)$ reshapes the token
sequence into a spatial latent map of resolution $H' \times W'$.

The model is trained using the v-prediction
objective~\cite{kingma2023understanding}:
\begin{equation}
\mathcal{L}_{\text{v-pred}}
= \mathbb{E}_t\!\left[\big\| \mathbf{w}_t \cdot
(\hat{\mathbf{v}}^{t} - (\boldsymbol{\epsilon} -
\mathbf{z}_{\text{img}})) \big\|_2^2 \right],
\end{equation}
where $\mathbf{w}_t$ is a step-dependent weighting term that balances
noise levels across timesteps. At inference, the same architecture
operates on VLM-derived $\mathbf{V}$ and $\mathbf{U}$, and the final
denoised latent is decoded through the VAE decoder to produce the
output image.

\section{Extended Implementation Details}
\label{sec:extended_imp}

\subsection{VQ-VAE Specifications}
Our default backbone, Janus-Pro 7B, employs a VQ-GAN-style VQ-VAE
as its image tokenizer for visual generation. The tokenizer encodes
each RGB image into an 8-channel latent representation with a
$16\times$ spatial downsampling factor, producing a $24 \times 24$
grid (576 spatial locations) for a $384 \times 384$ input
resolution. Each latent vector is vector-quantized by selecting the
nearest entry from a codebook of 16{,}384 learned embeddings, where
each code vector is 8-dimensional. During VLM-based image
generation, the multimodal transformer autoregressively predicts
576 discrete token IDs, each corresponding to one of the 16{,}384
codebook entries; the selected code vectors are arranged into an
8-channel $24 \times 24$ latent grid, which the VQ-VAE decoder maps
back to the RGB image space through learned dequantization and
upsampling layers.

\subsection{Decoder Training and Inference Hyperparameters}
We finetune the FLUX.1-Fill backbone with only its attention and
adapter layers updated; all other parameters are frozen. Training is
performed on the ImageNet-1K training set with standard resize and
center-crop preprocessing, using AdamW with learning rate
$2 \times 10^{-5}$, batch size 32, and 50{,}000 training steps on
$4 \times$ A100 80GB GPUs in mixed precision. At inference, the
decoder uses 30 denoising steps; VLM-side generation settings
($w_{\text{cfg}} = 5.0$, $\tau_{\text{VLM}} = 1.0$) match those used
during calibration (Appendix~\ref{sec:extended_lc}).

\section{Extended Conceptual Motivation}
\label{sec:appendix_preliminary}

This appendix provides the formal definition of Otsu's thresholding
used in Section~\ref{sec:motivation}, reports the post-calibration
analysis showing that Logit Calibration restores the encoder's
distributional sharpness to match the VLM, verifies position by
position that calibrated proxies and VLM logits provide compatible
decoder conditioning (Appendix~\ref{sec:proxy_validity}), and
analyzes the stability of that conditioning
(Appendix~\ref{sec:stability}).

\paragraph{Otsu's thresholding.} Otsu's method~\cite{otsu1979threshold} is a classical
variance-maximization technique that partitions a probability sequence
into statistically meaningful ``foreground'' and ``background''
components by selecting the threshold that maximizes inter-class
variance. Given sorted probabilities
$\{p_{(1)}, \dots, p_{(K)}\}$, the optimal threshold $\tau^\star$ is
obtained by
\begin{equation}
\begin{aligned}
\tau^\star
&= \arg\max_{\tau}\; \sigma^2_{\text{between}}(\tau) \\
&= \omega_0(\tau)\,\omega_1(\tau)\,
   \big( \mu_0(\tau) - \mu_1(\tau) \big)^2,
\end{aligned}
\end{equation}
where $\omega_c$ and $\mu_c$ denote the total mass and mean of each
partition, respectively. The corresponding threshold rank is the
number of probabilities at or above $\tau^\star$, indicating how many
codebook entries are statistically distinguishable from the noise
floor.

\paragraph{Effect of Logit Calibration on encoder distributions.}
Section~\ref{sec:motivation} establishes that raw proxy logits from
the VQ-VAE encoder are nearly uniform (Otsu rank $\approx 5{,}767$),
in stark contrast to the VLM's selective belief
(Otsu rank 2). Table~\ref{tab:otsu} reports the corresponding
statistics after applying our Logit Calibration. The calibration
substantially restores the desired sharpness: the Otsu threshold rank
becomes 2, identical to that of the VLM, and the top-1 probability
and head mass closely match the VLM's values. This demonstrates that
calibration reshapes the proxy-logit distribution into one that
faithfully reproduces the semantic selectivity exhibited by the VLM.

\begin{figure}[h]
  \centering
  \includegraphics[width=.7\linewidth]{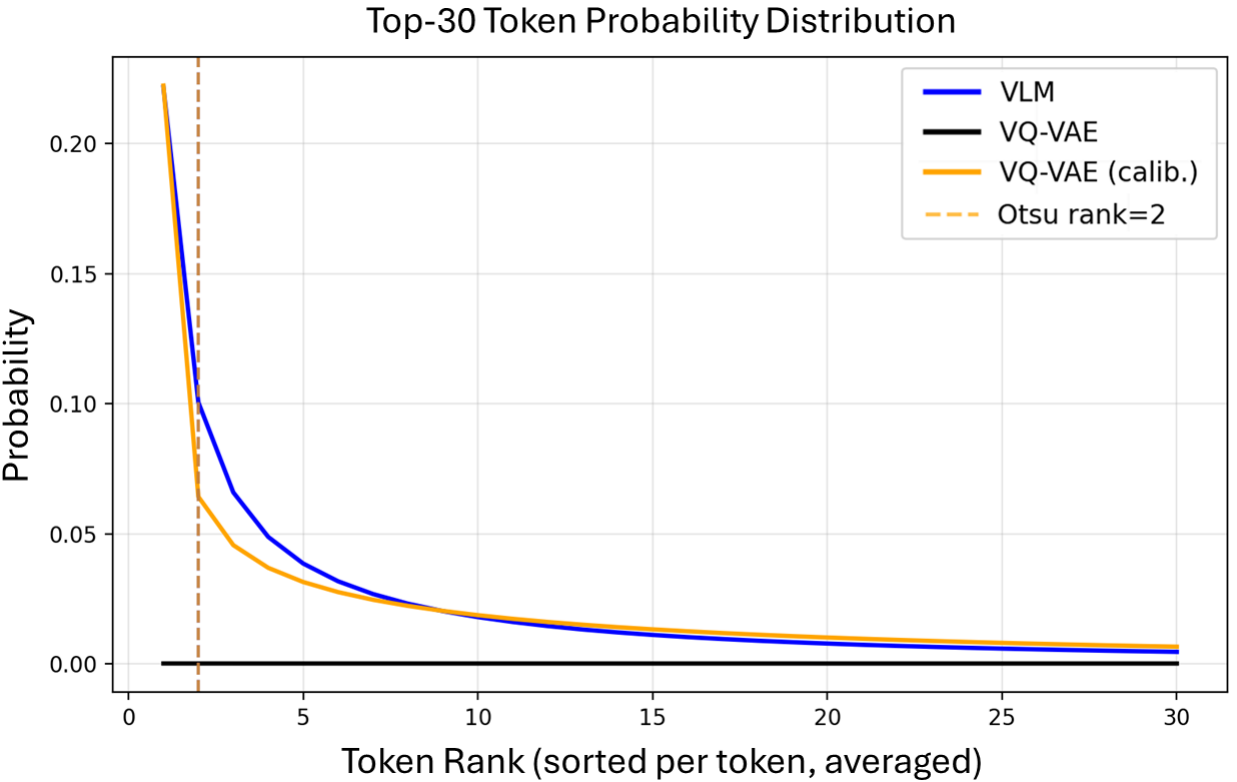}
  \caption{\textbf{Average top-30 token probability distributions} for
  Janus-Pro 7B (VLM), the raw VQ-VAE encoder, and the calibrated
  VQ-VAE encoder. Each curve shows per-token probabilities sorted by
  rank and averaged across 100 images. The dashed line marks the
  Otsu-derived threshold rank. The calibrated encoder closely follows
  the VLM's decay pattern, while the uncalibrated encoder produces a
  nearly flat distribution that no longer reflects semantic
  selectivity.}
  \label{fig:otsu}
\end{figure}

Figure~\ref{fig:otsu} visualizes the same effect across the full
top-30 distribution shape. The calibrated encoder distribution
closely tracks the VLM's decay across all visible ranks, whereas the
uncalibrated encoder remains nearly flat. This confirms that LC is
essential for enabling LCDM to operate on probability distributions
that faithfully reflect the VLM's semantic token structure, reducing
the discrepancy between training-time conditioning via the VQ-VAE
encoder and inference-time conditioning via the VLM.

\subsection{Correspondence between Calibrated Proxies and VLM Logits}
\label{sec:proxy_validity}

LC matches four global statistics, which alone do not guarantee that
calibrated proxies and VLM logits place mass on the same codebook
entries. We test this directly. For each of 200 MJHQ-30K Animals
prompts, disjoint from the 100 GenEval prompts used to fit LC, we
record the VLM's pre-sampling distribution and sampled token at all
576 positions, decode the sampled sequence with the native VQ-VAE
decoder, re-encode the resulting image with the same VQ-VAE encoder
used to construct training proxies, and form raw and calibrated
proxies. The round trip is faithful: 87.92\% of positions recover the
exact token, and the sampled token lies in the proxy's top-8 at
98.47\% of positions.

Since the 16{,}384-dimensional distribution never reaches the
decoder, we compare what the decoder actually receives: the
distribution-weighted code vector (Eq.~\eqref{eq:v}) and the four
uncertainty features (Eq.~\eqref{eq:uncertainty}) at each position.
As a control, we keep the calibrated proxy's probability values but
reassign them to permuted codebook entries, which preserves every
global statistic.

\begin{table}[h]
\centering
\caption{\textbf{Position-wise agreement between proxy and VLM
conditioning} (Janus-Pro 7B, 200 MJHQ-30K prompts, 576 positions
each). Permuting code identities preserves the uncertainty features
but destroys code-vector agreement.}
\resizebox{0.99\linewidth}{!}{
\begin{tabular}{lcc}
\Xhline{4\arrayrulewidth}
\rowcolor{table-yellow!60}
\textit{\textbf{Proxy vs.\ VLM logits}} & Code-vector cosine$\uparrow$ & Uncertainty MAE$\downarrow$ \\
\midrule
Raw proxy & 0.8125 & 0.4327 \\
Calibrated proxy & \hl{0.8647} & \hl{0.1220} \\
Calibrated proxy, code identities permuted & 0.0379 & 0.1220 \\
\Xhline{4\arrayrulewidth}
\end{tabular}
}
\label{tab:proxy_validity}
\end{table}

Permutation leaves the uncertainty features unchanged but collapses
the code-vector cosine from 0.8647 to 0.0379
(Table~\ref{tab:proxy_validity}). The agreement therefore does not
come from matched entropy and confidence alone; it comes from the
VQ-VAE encoder and the VLM placing mass on the same codebook entries
at the same position. LC then aligns their concentration: a positive
scale, a positive temperature, and uniform label smoothing cannot
reorder a distribution and only change how much weight each code
carries, reducing the uncertainty-feature error by 72\%.

\textbf{Source substitution.} We further test whether the two sources
induce similar decoder outputs. For the same 200 token sequences, we
decode with the same DCDD checkpoint while changing only the
conditioning source, using the same initial diffusion noise, and
compare against outputs conditioned on actual VLM logits. As a
reference for natural variation, we also change only the diffusion
noise.

\begin{table}[h]
\centering
\caption{\textbf{Source substitution at inference} (Janus-Pro 7B).
Similarity to the output conditioned on actual VLM logits (mean
$\pm$ SD over 200 prompts).}
\resizebox{0.99\linewidth}{!}{
\begin{tabular}{lccc}
\Xhline{4\arrayrulewidth}
\rowcolor{table-yellow!60}
\textit{\textbf{Conditioning source}} & CLIP-I$\uparrow$ & DINOv2$\uparrow$ & LPIPS$\downarrow$ \\
\midrule
Same source, different diffusion noise & 0.9539 $\pm$ 0.0265 & 0.9396 $\pm$ 0.0405 & 0.1045 $\pm$ 0.0328 \\
Calibrated encoder proxy & 0.9403 $\pm$ 0.0401 & 0.9088 $\pm$ 0.0692 & 0.1380 $\pm$ 0.0509 \\
\Xhline{4\arrayrulewidth}
\end{tabular}
}
\label{tab:source_sub}
\end{table}

Replacing VLM logits with the calibrated proxy changes the output only
slightly more than changing the diffusion noise
(Table~\ref{tab:source_sub}). We do not claim that the two raw
distributions are identical: the encoder observes a complete image,
whereas the VLM predicts causally from the prompt and previous
tokens. DCDD only requires them to induce compatible code-vector and
uncertainty conditioning. Note also that all text-to-image results in
the paper are computed with actual VLM logits; proxies are used only
during training.

\subsection{Stability of the Decoder Conditioning}
\label{sec:stability}

Matching four global statistics cannot guarantee equivalence between
two full distributions over $K = 16{,}384$ codes, and we do not claim
a generalization bound. Instead, we show that the quantities the
decoder receives change in a controlled way when the underlying
distribution changes. For distributions $\mathbf{p}, \mathbf{p}'$
over the codebook, let
$\delta = \mathrm{TV}(\mathbf{p}, \mathbf{p}') =
\tfrac{1}{2}\|\mathbf{p}-\mathbf{p}'\|_1$, and let
$R = \min_{\mathbf{c}} \max_k \|\mathbf{e}_k - \mathbf{c}\|_2$ be the
radius of the codebook.

\textbf{Proposition.} \textit{Let $(\mathbf{v}, \mathbf{u})$ and
$(\mathbf{v}', \mathbf{u}')$ be the LCDM outputs of $\mathbf{p}$ and
$\mathbf{p}'$. Then
(i) $\|\mathbf{v} - \mathbf{v}'\|_2 \le 2R\,\delta$;
(ii) $|u^{(1)} - u'^{(1)}| \le \delta$;
(iii) $|u^{(2)} - u'^{(2)}| \le 2\delta$; and
(iv) if both use the same $K_u$, $|u^{(3)} - u'^{(3)}| \le \delta$.}

\textit{Proof.} (i) Since $\sum_k (p_k - p'_k) = 0$, for any
$\mathbf{c}$ we have
$\mathbf{v} - \mathbf{v}' = \sum_k (p_k - p'_k)(\mathbf{e}_k - \mathbf{c})$,
so $\|\mathbf{v}-\mathbf{v}'\|_2 \le \max_k\|\mathbf{e}_k - \mathbf{c}\|_2\,
\|\mathbf{p}-\mathbf{p}'\|_1$; minimizing over $\mathbf{c}$ gives the
claim. (ii) For every $k$, $|p_k - p'_k| \le \delta$, and the maximum
is 1-Lipschitz in $\|\cdot\|_\infty$. (iii) Each order statistic
$p_{(j)}$ is 1-Lipschitz in $\|\cdot\|_\infty$, so the margin changes
by at most $2\|\mathbf{p}-\mathbf{p}'\|_\infty \le 2\delta$. (iv) The
top-$K_u$ mass equals $\max_{|A| = K_u} P(A)$, and
$|\max_A P(A) - \max_A P'(A)| \le \max_A |P(A) - P'(A)| \le \delta$.
$\square$

\textbf{Scope.} The tail entropy $u^{(4)}$ is continuous but not
Lipschitz near zero probabilities, and the entropy-adaptive $K_u$ is
piecewise constant, so (iv) holds whenever both distributions select
the same $K_u$; we therefore measure the uncertainty features
empirically (Table~\ref{tab:proxy_validity}). The proposition bounds
the change in conditioning, not the distance $\delta$ between proxy
and VLM distributions, nor the response of the decoder to a change in
conditioning. In addition, the LC transform is order-preserving: for
$a > 0$, $\alpha > 0$, and $\epsilon \in [0, 1)$,
$\hat p_{i,k} > \hat p_{i,k'}$ if and only if $s_{i,k} > s_{i,k'}$.
Calibration therefore changes only the concentration of each proxy
distribution and leaves its ranking of codebook entries, and hence
its top-$k$ sets, unchanged; the agreement of those rankings with the
VLM is an empirical property, measured in
Appendix~\ref{sec:proxy_validity}.

\section{Additional Ablation Studies}
\label{sec:appendix_ablation}

\subsection{Denoising Steps}
\label{sec:appendix_denoisingsteps}
We evaluate our diffusion decoder under different numbers of
denoising steps while keeping all other components fixed, measuring
reconstruction quality on the first 1{,}000 validation images of
ImageNet-1K against the VQ-VAE decoder of Janus-Pro 7B.
Table~\ref{tab:steps} summarizes the results.

\begin{table}[h]
    \centering
    \caption{\textbf{Reconstruction performance across denoising steps}
    (Janus-Pro 7B, ImageNet-1K validation). All rows use the final
    50K-step decoder; absolute rFID values are higher than those in
    Table~\ref{tab:recon} because evaluation here uses a 1{,}000-image
    subset rather than the full validation split. Best results are
    \hl{highlighted}.}
    \resizebox{0.48\textwidth}{!}{
    \begin{tabular}{lccc}
    \Xhline{4\arrayrulewidth}
        \rowcolor{table-yellow!60}\textit{\textbf{ImageNet-1K Val.}} & rFID$\downarrow$ & SSIM$\uparrow$ & LPIPS$\downarrow$ \\
        \midrule
        VQ-VAE Decoder & 23.1038 & 0.5756 & 0.1551 \\
        \midrule
        Ours (5 steps) & 35.0169 & 0.5855 & 0.1918 \\
        Ours (10 steps) & 20.8859 & 0.5995 & 0.1478 \\
        Ours (15 steps) & 18.3279 & \hl{0.6012} & 0.1417 \\
        Ours (20 steps) & 17.6918 & 0.5993 & 0.1405 \\
        Ours (25 steps) & \hl{17.5659} & 0.5975 & \hl{0.1404} \\
        Ours (30 steps) & 17.6969 & 0.5954 & 0.1410 \\
        Ours (35 steps) & 17.7818 & 0.5940 & 0.1414 \\
    \Xhline{4\arrayrulewidth}
    \end{tabular}
    }
    \label{tab:steps}
\end{table}

Even with 10 denoising steps, our decoder surpasses the VQ-VAE
baseline across all metrics. As the step count increases, rFID and
LPIPS reach their best values at 25 steps before gradually
degrading, while SSIM peaks earlier at 15 steps, suggesting that
global structure is recovered in the initial denoising stages and
subsequent steps refine texture and low-frequency detail. The
trajectory thus rapidly establishes the core scene layout, with
additional steps primarily polishing perceptual details rather than
altering structural content.

\paragraph{Choice of inference denoising steps.}
Although 25 steps yields the lowest rFID and LPIPS, the gap to 30
steps is marginal in absolute terms (rFID 17.57 vs 17.70, LPIPS
0.1404 vs 0.1410, SSIM 0.5975 vs 0.5954). We adopt 30 denoising
steps throughout the paper, matching the default sampling schedule
of FLUX.1-Fill, our diffusion backbone. This choice avoids
backbone-specific hyperparameter tuning and provides a more
reproducible setting: practitioners finetuning FLUX.1-Fill on
related tasks can use the model's standard sampling configuration
without re-running this ablation on their setup.

\subsection{Latency and Memory}
\label{sec:latency}

Table~\ref{tab:latency} reports per-image latency across denoising
steps, measured in BF16 at $384 \times 384$ with batch size 1, two
warm-up runs, and seven synchronized runs on an A100 80GB. Denoising
cost is linear at approximately 0.12\,s per step, and VLM
autoregression takes 11.84\,s. Ten steps already outperform the
native decoder at 10.7\% additional end-to-end latency, and 15 steps
recover 86\% of the best fidelity gain using 51\% of the additional
latency of the 30-step setting.

\begin{table}[h]
\centering
\caption{\textbf{Latency across denoising steps} (Janus-Pro 7B, A100
80GB, batch size 1). rFID is measured on the 1{,}000-image
ImageNet-1K subset of Table~\ref{tab:steps}.}
\resizebox{0.99\linewidth}{!}{
\begin{tabular}{lcccc}
\Xhline{4\arrayrulewidth}
\rowcolor{table-yellow!60}
\textit{\textbf{Decoder}} & Steps & Decoder-only (s/img)$\downarrow$ & End-to-end (s/img)$\downarrow$ & rFID$\downarrow$ \\
\midrule
VQ-VAE Decoder & -- & 0.02 & 11.86 & 23.1038 \\
\midrule
DCDD & 10 & 1.10 & 13.13 & 20.8859 \\
DCDD & 15 & 1.71 & 13.74 & 18.3279 \\
DCDD & 20 & 2.32 & 14.35 & 17.6918 \\
DCDD & 25 & 2.93 & 14.96 & \hl{17.5659} \\
DCDD & 30 & 3.53 & 15.56 & 17.6969 \\
\Xhline{4\arrayrulewidth}
\end{tabular}
}
\label{tab:latency}
\end{table}

Peak allocated memory is 14.90\,GiB for the VLM, 0.32\,GiB for the
native decoder, and 22.67\,GiB for the DCDD decoder, or approximately
37.6\,GiB for the full pipeline. This exceeds typical
edge-device budgets; deployment in that setting would require a
quantized or distilled decoder (Appendix~\ref{sec:limitations}).

\subsection{Training Dynamics of Component Ablations}
\label{sec:stepwise}

Table~\ref{tab:stepwise} tracks three decoders that differ only in
their conditioning across training checkpoints up to 10K steps:
(i) discrete code vectors; (ii) DCDD without LC, with continuous
distribution-weighted code vectors and uncertainty features but no
calibration; and (iii) full DCDD. LCDM provides the largest single
rFID reduction, LC provides the largest single SSIM improvement, and
LPIPS decreases monotonically across the three variants. The relative
ranking emerges by 5K steps and remains stable, indicating that the
contributions are driven by the components rather than by
training-time noise. The argmax and full variants are compared at the
full 50K-step schedule in Table~\ref{tab:argmax}.

\begin{table}[h]
\centering
\caption{\textbf{Training-step ablation on ImageNet-1K
reconstruction} (Janus-Pro 7B). We track three variants up to 10K
training steps under matched optimization: \textit{Diff.\ Dec.\
(baseline)}, with discrete code-vector conditioning; \textit{DCDD
w/o LC}, with continuous distribution-weighted code vectors and
uncertainty features but no calibration; and \textit{DCDD (full)},
with calibration applied to the proxy logits. The relative ranking
among variants emerges by 5K steps and remains stable.
Best value at each step for each metric is \hl{highlighted}.
}
\resizebox{\textwidth}{!}{
\begin{tabular}{cccccccccc}
\Xhline{4\arrayrulewidth}
\rowcolor{table-yellow!60}
& \multicolumn{3}{c}{\textbf{\textit{Diff.\ Dec.\ (baseline)}}}
& \multicolumn{3}{c}{\textbf{\textit{DCDD w/o LC}}}
& \multicolumn{3}{c}{\textbf{\textit{DCDD (full)}}} \\
\cline{2-10}
\rowcolor{table-yellow!60}
\textbf{Steps}
& rFID $\downarrow$ & SSIM $\uparrow$ & LPIPS $\downarrow$
& rFID $\downarrow$ & SSIM $\uparrow$ & LPIPS $\downarrow$
& rFID $\downarrow$ & SSIM $\uparrow$ & LPIPS $\downarrow$ \\
\Xhline{2\arrayrulewidth}
2.5K
& 173.7125 & 0.1940 & 0.7821
& 161.4700 & 0.1949 & \hl{0.7732}
& \hl{158.7311} & \hl{0.1965} & 0.7801 \\
5.0K
& 106.0594 & 0.2279 & 0.6359
& 93.4573 & 0.2717 & 0.6051
& \hl{91.8926} & \hl{0.2750} & \hl{0.5833} \\
7.5K
& 45.3929 & 0.3080 & 0.5713
& 39.5000 & 0.3147 & 0.5558
& \hl{37.8744} & \hl{0.3222} & \hl{0.5498} \\
10.0K
& 33.5081 & 0.4244 & 0.4132
& 31.2049 & 0.4275 & 0.4009
& \hl{30.5760} & \hl{0.4360} & \hl{0.3875} \\
\Xhline{3\arrayrulewidth}
\end{tabular}
}
\label{tab:stepwise}
    \vskip -0.1in
\end{table}

\subsection{Comparison of Post-hoc Alternatives}
\label{sec:argmax_full}

Table~\ref{tab:posthoc} places the matched argmax-conditioned
diffusion decoder of Section~\ref{sec:cctd} alongside the
image-to-image refiners of Section~\ref{sec:llcvspsr}, with the VLM
and tokenizer fixed. The argmax decoder attains the highest CLIP-I and
DINOv2 similarity because its conditioning is exactly the code
sequence that the VQ-VAE reference image is decoded from, so it stays
closest to that reference by construction. DCDD departs slightly
further from the reference as it uses the additional distributional
information to recover detail, which is reflected in the lower gFID,
while remaining well above both I2I refiners in similarity.

\begin{table}[h]
\centering
\caption{\textbf{Post-hoc alternatives on MJHQ-30K} (Janus-Pro 7B,
VLM and tokenizer fixed). CLIP-I and DINOv2 are measured against the
VQ-VAE-decoded image.}
\resizebox{0.65\linewidth}{!}{
\begin{tabular}{lccc}
\Xhline{4\arrayrulewidth}
\rowcolor{table-yellow!60}
\textit{\textbf{Method}} & gFID$\downarrow$ & CLIP-I$\uparrow$ & DINOv2$\uparrow$ \\
\midrule
VQ-VAE Decoder & 14.2888 & -- & -- \\
\textit{I2I (generic)} & 10.6678 & 0.8752 & 0.7663 \\
\textit{I2I + prompt} & 10.9767 & 0.8531 & 0.7586 \\
Diff.\ Dec.\ (argmax codes) & 11.0837 & 0.9275 & 0.8933 \\
\textbf{DCDD (ours)} & \hl{9.4160} & 0.9088 & 0.8692 \\
\Xhline{4\arrayrulewidth}
\end{tabular}
}
\label{tab:posthoc}
\end{table}

\subsection{Soft-Token Adapter without Diffusion}
\label{sec:adapter}

To test whether the distributional signal can be exploited without a
diffusion decoder, we train a token-wise residual MLP
($8 \to 64 \to 8$, 1{,}096 parameters, zero-initialized) placed
before the frozen native VQ-VAE decoder. Argmax and soft variants use
identical architecture, data, and optimization, with an L1 $+$ 0.5
LPIPS objective and code-displacement and norm penalties.
Table~\ref{tab:adapter} reports reconstruction on the first 1{,}000
ImageNet-1K validation images and generation on the MJHQ-30K Animals
category.

\begin{table}[h]
\centering
\caption{\textbf{Soft-token adapter before the frozen native decoder}
(Janus-Pro 7B). \textit{Top:} reconstruction (ImageNet-1K, 1{,}000
images). \textit{Bottom:} generation (MJHQ-30K Animals gFID and
GenEval).}
\resizebox{0.8\linewidth}{!}{
\begin{tabular}{lccccc}
\Xhline{4\arrayrulewidth}
\rowcolor{table-yellow!60}
\textit{\textbf{Reconstruction}} & Train steps & MSE$\downarrow$ & rFID$\downarrow$ & SSIM$\uparrow$ & LPIPS$\downarrow$ \\
\midrule
Argmax, no adapter & -- & 0.0360 & 23.10 & 0.5756 & 0.1551 \\
Expected code, no adapter & 0 & 0.0285 & 22.71 & 0.6081 & 0.1418 \\
Adapter on argmax & 100 & 0.0345 & 23.00 & 0.5829 & 0.1550 \\
Adapter on expected code & 2K & \hl{0.0274} & \hl{22.26} & \hl{0.6164} & \hl{0.1390} \\
\Xhline{4\arrayrulewidth}
\end{tabular}
}
\resizebox{0.6\linewidth}{!}{
\begin{tabular}{lcc}
\Xhline{3\arrayrulewidth}
\rowcolor{table-yellow!60}
\textit{\textbf{Generation}} & gFID$\downarrow$ (Animals) & GenEval$\uparrow$ \\
\midrule
VQ-VAE Decoder & 31.68 & 77.24 \\
+ Adapter on argmax & 31.81 & 77.12 \\
+ Adapter on expected code & 34.28 & 75.41 \\
\textbf{DCDD (ours)} & \hl{25.10} & \hl{78.31} \\
\Xhline{4\arrayrulewidth}
\end{tabular}
}
\label{tab:adapter}
\end{table}

Expected codes improve all reconstruction metrics with no trained
parameters, and the soft adapter improves them further, so the
distributional signal is useful independently of diffusion. In
generation, however, the soft adapter is worse than the native
decoder, while the argmax adapter is indistinguishable from it. In
reconstruction the expected codes come from the VQ-VAE encoder,
whereas in generation they come from the VLM; a frozen decoder
trained on quantized codes cannot learn to read a distribution
through a small residual. Exploiting the signal for generation
therefore requires a decoder trained for distributional
conditioning.

\subsection{Variation across Diffusion-Noise Seeds}
\label{sec:seeds}

We re-run DCDD with two additional diffusion-noise seeds, reusing the
same VLM token sequences so that only the initial diffusion noise
changes. Table~\ref{tab:seeds} reports the mean and standard
deviation over the three runs for the overall metrics of
Tables~\ref{tab:recon}, \ref{tab:t2i}, and~\ref{tab:i2i}. All values
reported in the main tables fall within the observed range.

\begin{table}[h]
\centering
\caption{\textbf{DCDD across three diffusion-noise seeds} (Janus-Pro
7B, mean $\pm$ SD; VLM token sequences fixed).}
\resizebox{0.6\linewidth}{!}{
\begin{tabular}{llc}
\Xhline{4\arrayrulewidth}
\rowcolor{table-yellow!60}
\textit{\textbf{Table}} & \textit{\textbf{Metric}} & Mean $\pm$ SD \\
\midrule
\multirow{6}{*}{Table~\ref{tab:recon}} & ImageNet-1K rFID$\downarrow$ & 0.7671 $\pm$ 0.0478 \\
& ImageNet-1K SSIM$\uparrow$ & 0.5957 $\pm$ $<$0.0001 \\
& ImageNet-1K LPIPS$\downarrow$ & 0.1374 $\pm$ 0.0001 \\
& MJHQ-30K rFID$\downarrow$ & 1.8174 $\pm$ 0.0480 \\
& MJHQ-30K SSIM$\uparrow$ & 0.6782 $\pm$ $<$0.0001 \\
& MJHQ-30K LPIPS$\downarrow$ & 0.1307 $\pm$ 0.0001 \\
\midrule
\multirow{4}{*}{Table~\ref{tab:t2i}} & MJHQ-30K gFID$\downarrow$ & 9.3787 $\pm$ 0.2929 \\
& GenEval Overall$\uparrow$ & 78.32 $\pm$ 0.10 \\
& DPG-Bench Overall$\uparrow$ & 87.64 $\pm$ 0.03 \\
& WISE Consistency$\uparrow$ & 1.318 $\pm$ 0.012 \\
\midrule
\multirow{2}{*}{Table~\ref{tab:i2i}} & CLIP-I$\uparrow$ & 0.9085 $\pm$ 0.0022 \\
& DINOv2$\uparrow$ & 0.8701 $\pm$ 0.0036 \\
\Xhline{4\arrayrulewidth}
\end{tabular}
}
\label{tab:seeds}
\end{table}

\section{Sensitivity of Logit Calibration to Reference Prompts}
\label{sec:lc_sensitivity}

\begin{figure}[h]
  \centering
  \includegraphics[width=1.0\linewidth]{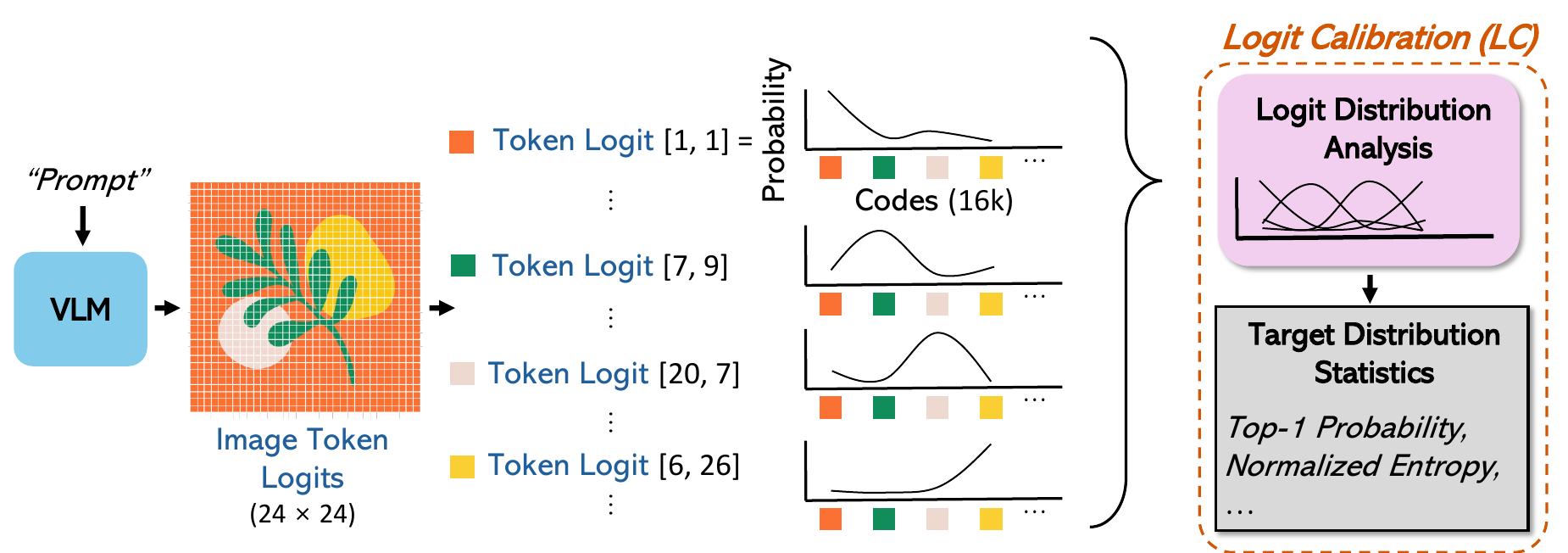}
  \caption{\textbf{Logit Calibration matches coarse global statistics
  of VLM logits.} Each $384 \times 384$ image produces a $24 \times
  24$ grid of image-token logits, i.e., 576 patch-level codebook
  logit distributions. LC analyzes these VLM logits to estimate
  global statistics such as confidence and entropy, which are then
  used as calibration targets for the proxy logits derived from the
  VQ-VAE encoder.}
  \label{fig:logit_calib}
\end{figure}

Logit Calibration estimates four target statistics from a sample of
VLM-generated images and uses them to align proxy-logit
distributions to the VLM's regime. Two natural concerns about this
setup are: (i) whether the sample is large enough for the estimates
to be stable, and (ii) whether the choice of prompt set biases the
resulting calibration toward a particular benchmark's content.
Figure~\ref{fig:logit_calib} illustrates the calibration data flow.
This appendix addresses both concerns.

\subsection{Sensitivity to Sample Size}
\label{sec:lc_samplesize}

We vary the number of reference prompts from 1 to 200 and measure
the resulting global logit scale $a$ alongside the target statistics
it is fit against. We focus on $a$ because it is the primary
calibration parameter, fit first against the VLM's mean entropy in
our gradient-free search; the remaining parameters
$(b, \alpha, \epsilon)$ are estimated subsequently and are
conditioned on the choice of $a$. Table~\ref{tab:lc_samplesize}
reports the results.

\begin{table}[h]
    \centering
    \caption{\textbf{Sensitivity of Logit Calibration to the number
    of reference prompts} (Janus-Pro 7B). The estimated global logit
    scale $a$ and target statistics (mean normalized entropy, mean
    confidence) remain within a narrow range across sample sizes,
    with no clear trend as the number of prompts grows. Each prompt
    contributes 576 token-level distributions, so even a modest
    number of prompts provides sufficient coverage to estimate
    stable global statistics.}
    \resizebox{0.65\textwidth}{!}{
    \begin{tabular}{c|c||cc}
    \Xhline{4\arrayrulewidth}
        \rowcolor{table-yellow!60}
        \textit{\textbf{\# Prompts}} &
        \textit{\textbf{$a$}} &
        \textit{\textbf{Mean Entropy}} &
        \textit{\textbf{Mean Confidence}} \\
        \midrule
        1   & 28.4570 & 0.4318 & 0.1807 \\
        5   & 30.2344 & 0.4123 & 0.1887 \\
        10  & 29.5679 & 0.4196 & 0.1846 \\
        20  & 28.3716 & 0.4328 & 0.1743 \\
        50  & 28.5938 & 0.4302 & 0.1766 \\
        100 & 29.0381 & 0.4254 & 0.1797 \\
        150 & 29.0894 & 0.4248 & 0.1801 \\
        200 & 29.0210 & 0.4255 & 0.1797 \\
    \Xhline{4\arrayrulewidth}
    \end{tabular}
    }
    \label{tab:lc_samplesize}
\end{table}

The estimated $a$ stabilizes from 100 prompts onward (29.04, 29.09,
29.02 at 100, 150, and 200 prompts respectively), with mean entropy
in $[0.412, 0.432]$ and mean confidence in $[0.174, 0.189]$ across
the full range. The estimates show no monotonic trend with sample
size, indicating that target statistics primarily reflect stable,
model-level logit behavior rather than prompt-specific variation.
Our default choice of 100 prompts is at this stable regime.

\subsection{Sensitivity to Prompt Type}
\label{sec:lc_prompttype}

A separate question is whether calibration is sensitive to what the
reference prompts are, not how many. This bears on the concern that
calibrating on GenEval prompts could implicitly bias the decoder
toward GenEval-style content. We measure target statistics across
several qualitatively different prompt sets:

\begin{itemize}
\item \textbf{Natural-language benchmarks.} GenEval and MJHQ-30K
prompts (average prompt lengths 32.2 and 185.7 characters,
respectively).
\item \textbf{Constructed prompts.} Three categories synthesized to
stress the VLM in specific ways, all controlled to roughly 50
characters: \textit{(1) Unnatural} (lexically implausible),
\textit{(2) Compositional overload} (very dense object
specifications), and \textit{(3) Self-contradiction} (prompts
containing internal inconsistencies).
\item \textbf{Length-controlled random tokens.} Empty prompts and
random alphanumeric-with-space strings of varying lengths (50 to
1000 characters), which probe whether prompt length alone affects
the VLM's logit regime in the absence of semantic content.
\end{itemize}

\begin{figure}[h]
  \centering
  \includegraphics[width=1.0\linewidth]{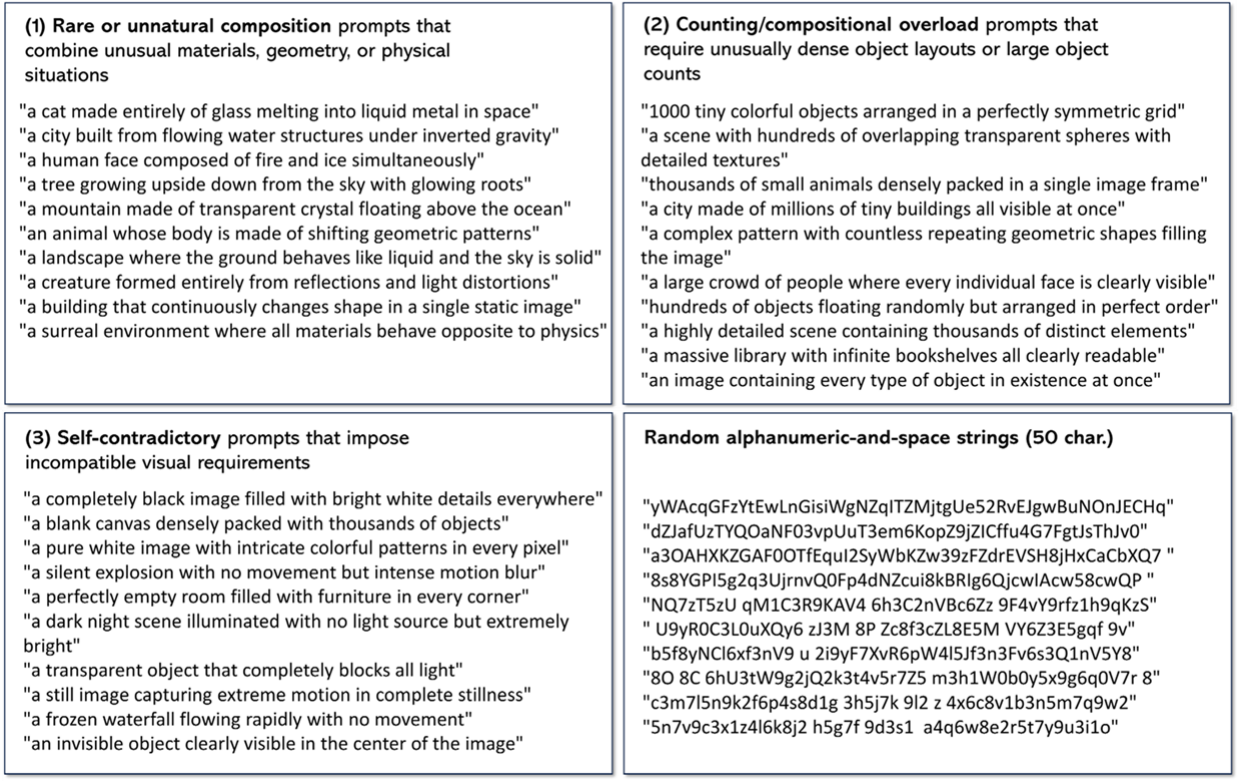}
  \caption{\textbf{Sample prompts from each prompt set used in the
  prompt-type analysis.} Three constructed prompt categories
  (Unnatural, Compositional overload, Self-contradiction) and
  length-controlled random alphanumeric-with-space prompts.}
  \label{fig:prompt_types}
\end{figure}

For a fair comparison, each prompt type is evaluated using 10
prompts. Sample prompts from each set are shown in
Figure~\ref{fig:prompt_types}; results are in
Table~\ref{tab:lc_prompttype}.

\begin{table}[h]
    \centering
    \caption{\textbf{Logit statistics across different prompt types}
    (Janus-Pro 7B, 10 prompts per type). Key values discussed in the
    text are \hl{highlighted}.}
    \resizebox{0.65\textwidth}{!}{
    \begin{tabular}{c|c||cc}
    \Xhline{4\arrayrulewidth}
        \rowcolor{table-yellow!60}
        \textit{\textbf{Prompt type}} &
        \textit{\textbf{$a$}} &
        \textit{\textbf{Mean Entropy}} &
        \textit{\textbf{Mean Confidence}} \\
        \midrule
        GenEval & \hl{29.5679} & 0.4195 & 0.1846 \\
        MJHQ-30K & \hl{32.1826} & 0.3920 & 0.2127 \\
        \midrule
        (1) Unnatural & 30.2856 & 0.4117 & 0.1901 \\
        (2) Comp. overload & \hl{25.8081} & 0.4632 & 0.1541 \\
        (3) Self-contradict. & \hl{24.4751} & 0.4800 & 0.1368 \\
        \midrule
        Blank prompt, ``'' & 27.2949 & 0.4452 & 0.1639 \\
        Rand. char. 50 & \hl{27.7905} & 0.4395 & 0.1645 \\
        Rand. char. 100 & 29.0381 & 0.4254 & 0.1821 \\
        Rand. char. 200 & \hl{30.6104} & 0.4083 & 0.1986 \\
        Rand. char. 300 & 31.7383 & 0.3966 & 0.2100 \\
        Rand. char. 400 & 30.7983 & 0.4064 & 0.2018 \\
        Rand. char. 500 & 29.3799 & 0.4215 & 0.1866 \\
        Rand. char. 1000 & 29.5166 & 0.4201 & 0.1894 \\
    \Xhline{4\arrayrulewidth}
    \end{tabular}
    }
    \label{tab:lc_prompttype}
\end{table}

The data reveal two effects that disentangle semantic content from
prompt length.

\textbf{Semantically extreme prompts increase uncertainty, not
decrease it.} The constructed Compositional-overload and
Self-contradiction prompts produce noticeably lower $a$ (25.8 and
24.5) and higher entropy than GenEval, consistent with the VLM
having more ambiguity about what to render. However, MJHQ-30K
produces \emph{higher} $a$ (32.2) than GenEval, in the opposite
direction. If semantic content drove the natural-prompt gap, we
would expect MJHQ-30K (a more diverse, free-form benchmark) to
produce higher uncertainty, not lower. Instead, MJHQ-30K is sharper
than GenEval, ruling out semantic content as the dominant driver.

\textbf{Length-controlled random prompts reproduce the natural-prompt
gap.} Random alphanumeric-and-space strings strip out semantic
content entirely while varying length. Across these length-only
controls, $a$ rises from 27.8 (50 characters) to 30.6 (200
characters), a difference of 2.8. The gap between GenEval (about 30 characters,
$a = 29.6$) and MJHQ-30K (about 180 characters, $a = 32.2$) is 2.6, comparable
in magnitude. Since random prompts contain no benchmark-specific
content, the comparable gap suggests that the GenEval-MJHQ-30K
difference is driven primarily by prompt length rather than by
GenEval-specific or MJHQ-specific semantic structure.

These two observations together address the concern that calibrating
on GenEval introduces benchmark-specific bias. The calibration
target is governed by length-dependent properties of the VLM's logit
regime that hold across diverse prompt content; GenEval prompts are
neither outliers nor uniquely informative for the calibration. We
also note that LCDM accounts for the majority of the
fidelity gain (Appendix~\ref{sec:stepwise}), so even if the calibrated scale of the LC shifted modestly under different reference prompts, the downstream impact on decoder performance would be minor at this scale of difference. 

\section{Regime of Applicability: Results on LlamaGen-XL}
\label{sec:appendix_vlm}

A post-hoc decoder cannot recover information absent from its input
token sequence. As summarized in Section~\ref{sec:regime}, DCDD's
effectiveness depends on the upstream model's logit structure; here we characterize that
dependence empirically by evaluating DCDD on a backbone outside the
regime where the main paper's quantitative claims hold:
LlamaGen-XL at $256 \times 256$ resolution, with baseline T2I
performance of GenEval 30.0 (vs.\ Janus-Pro 7B's 77.2) and MJHQ-30K
gFID 59.0 (vs.\ 14.3). The two backbones thus span a wide range of
upstream token quality, allowing us to characterize the regime in
which post-hoc decoding is effective.

\subsection{Image Reconstruction}
\label{sec:llamagen_recon}

We first evaluate reconstruction, where the token sequence comes
from the VQ-VAE encoder and bypasses the VLM entirely. Quantitative
results are in Table~\ref{tab:recon_llamagen}, with qualitative
samples in Figure~\ref{fig:recon_llamagen}.

\begin{table}[h]
    \centering
    \caption{\textbf{Reconstruction performance on LlamaGen-XL}
    ($256 \times 256$). Best results per metric are
    \hl{highlighted}.}
    \resizebox{0.5\textwidth}{!}{
    \begin{tabular}{lccc}
    \Xhline{4\arrayrulewidth}
        \rowcolor{table-yellow!60}\textit{\textbf{ImageNet-1K Val.}} & rFID$\downarrow$ & SSIM$\uparrow$ & LPIPS$\downarrow$ \\
        \midrule
        VQ-VAE Decoder & 4.1919 & 0.5566 & \hl{0.1388} \\
        \textbf{DCDD (ours)} & \hl{1.4087} & \hl{0.5600} & 0.1404 \\
    \end{tabular}
    }
    \resizebox{0.5\textwidth}{!}{
    \begin{tabular}{lccc}
    \Xhline{3\arrayrulewidth}
        \rowcolor{table-yellow!60}\textit{\textbf{MJHQ-30K}} & rFID$\downarrow$ & SSIM$\uparrow$ & LPIPS$\downarrow$ \\
        \midrule
        VQ-VAE Decoder & 4.9653 & \hl{0.6140} & \hl{0.1407} \\
        \textbf{DCDD (ours)} & \hl{1.9028} & 0.6060 & 0.1484 \\
    \Xhline{4\arrayrulewidth}
    \end{tabular}
    }
    \label{tab:recon_llamagen}
\end{table}

DCDD substantially reduces rFID on both datasets (66\% on
ImageNet-1K, 62\% on MJHQ-30K), with SSIM and LPIPS remaining
comparable to the VQ-VAE baseline. The lower input resolution
provides fewer spatial samples for the diffusion decoder than the
$384 \times 384$ Janus-Pro setting, yet substantial reconstruction
gains still hold. 
This confirms that the decoder is not
backbone-specific for reconstruction: when the upstream token sequence carries
faithful information about the input image (as in reconstruction),
DCDD recovers fidelity across backbones.

\subsection{Text-to-Image Generation}
\label{sec:llamagen_t2i}

The result differs for end-to-end text-to-image generation, where
the token sequence is produced by the VLM from a text prompt
rather than from an image. Table~\ref{tab:t2i_llamagen} reports the
results.

\begin{table}[h]
\centering
\caption{\textbf{Text-to-image generation on LlamaGen-XL}
($256 \times 256$). DCDD inherits and amplifies the weaknesses of
the upstream token sequence on this backbone. Best results per
metric are \hl{highlighted}.}
\resizebox{1.0\textwidth}{!}{
    \begin{tabular}{lccccccccccc}
    \Xhline{4\arrayrulewidth}
    \multicolumn{12}{c}{\textbf{\textit{Fidelity Metric $\downarrow$}}} \\
    \Xhline{2\arrayrulewidth}
     \rowcolor{table-yellow!60}\textit{\textbf{MJHQ-30K}} & \textbf{Overall} & Animals & Art & Fashion & Food & Indoor & Landscape & Logo & People & Plants & Vehicles \\
    \Xhline{2\arrayrulewidth}
        VQ-VAE Decoder & \hl{59.0292} & \hl{74.1959} & \hl{82.9698} & \hl{92.8457} & \hl{86.4562} & \hl{89.2189} & \hl{82.0326} & 68.5263 & \hl{105.9981} & \hl{87.2854} & \hl{82.0326} \\
        \textbf{DCDD (ours)} & 61.0925 & 74.6621 & 87.1321 & 104.2776 & 93.6168 & 96.6309 & 90.0496 & \hl{66.9729} & 114.4049 & 93.7683 & 90.0496 \\
        \Xhline{3\arrayrulewidth}
    \end{tabular}
    }
    \resizebox{\linewidth}{!}{
    \begin{tabular}{lccccccc}
    \Xhline{4\arrayrulewidth}
    \multicolumn{8}{c}{\textbf{\textit{Semantic Consistency Checks}}} \\
    \Xhline{2\arrayrulewidth}
    \rowcolor{table-yellow!60}
    \textit{\textbf{GenEval}} & \textbf{Overall}
    & Single & Two & Counting
    & Colors  & Position & Color Attr. \\
    \Xhline{2\arrayrulewidth}
    VQ-VAE Decoder & \hl{30.02} & 69.06 & \hl{32.32} & 14.69 & \hl{55.85} & \hl{5.25} & \hl{3.00} \\
    \textbf{DCDD (ours)}
    & 28.19  & \hl{70.94}  & 25.76  & \hl{19.38} & 47.07 & 4.25 & 1.75 \\
    \Xhline{4\arrayrulewidth}
    \end{tabular}
    }
\label{tab:t2i_llamagen}
\end{table}

In contrast to the strong reconstruction gains, DCDD's T2I metrics
on LlamaGen-XL are essentially flat or mildly negative: gFID rises
slightly (59.03 to 61.09 on MJHQ-30K) and GenEval shifts marginally
(30.02 to 28.19), with sub-category movements scattered in both
directions. This is the opposite pattern from Janus-Pro 7B, where
DCDD reduces gFID by 34\% with stable semantics.

\paragraph{Why DCDD does not help on LlamaGen-XL.}
The difference traces to two factors. First, LlamaGen-XL's baseline
T2I quality is much weaker than Janus-Pro 7B's, indicating that the
underlying token sequence already contains substantial semantic and
perceptual errors. DCDD is designed to render the VLM's intended
output faithfully, not to correct semantic errors; when the
upstream token sequence is already semantically degraded, faithful
decoding inherits the degradation rather than fixing it.

Second, LlamaGen-XL's logit distributions leave less distributional
information for DCDD to exploit. Table~\ref{tab:otsu_comp} compares
the two backbones' image-token logit statistics. LlamaGen-XL is
more concentrated than Janus-Pro 7B: top-1 probability 0.38 vs.\
0.22, normalized entropy 0.33 vs.\ 0.42, and crucially, an Otsu
threshold rank of 1 vs.\ 2. Each LlamaGen-XL token already commits
strongly to a single codebook entry, leaving little rank-2 or
higher distributional information for LCDM to convert into
conditioning. This is the precondition our method relies on, as
established in Section~\ref{sec:motivation}: when the pre-argmax
logits do not carry meaningful rank-2 mass, there is little
discarded signal for a post-hoc decoder to recover.

\begin{table}[h]
\centering
\caption{\textbf{Token-probability statistics and Otsu thresholding
for Janus-Pro 7B vs.\ LlamaGen-XL.} LlamaGen-XL's distributions
already commit more strongly to a single codebook entry (top-1
0.38, Otsu rank 1) than Janus-Pro 7B's (top-1 0.22, Otsu rank 2),
leaving less distributional information beyond the top-1 entry for
LCDM to exploit.}
\resizebox{0.65\linewidth}{!}{
\begin{tabular}{lcc}
\Xhline{4\arrayrulewidth}
\rowcolor{table-yellow!60}
\textbf{Statistic} &
\textbf{Janus-Pro 7B} &
\textbf{LlamaGen-XL} \\
\midrule
\multicolumn{3}{l}{\textbf{\textit{Probability Distribution Statistics}}} \\
\midrule
Top-1 Probability      & 0.221627 & 0.379384  \\
Top-2 Probability      & 0.100532 & 0.117179 \\
Normalized Entropy     & 0.415303 & 0.327751 \\
Tail Entropy           & 0.560794 & 0.539393  \\
\midrule
\multicolumn{3}{l}{\textbf{\textit{Otsu Thresholding Statistics}}} \\
\midrule
Threshold (prob.)      & 6.24$\times10^{-2}$ & 1.17$\times10^{-1}$  \\
Threshold Rank         & 2        & 1      \\
Head Mass              & 0.322160 & 0.379384  \\
\Xhline{4\arrayrulewidth}
\end{tabular}
}
\label{tab:otsu_comp}
\end{table}

Together, these factors describe a regime of applicability:
post-hoc fidelity decoding via DCDD is most effective when (a) the
upstream VLM produces token sequences that already carry coherent
semantics, and (b) the per-token logit distribution retains
non-trivial rank-2 or higher mass that LCDM can convert into
continuous conditioning. Reconstruction satisfies both conditions
on both backbones. T2I generation on Janus-Pro 7B satisfies both
conditions; T2I generation on LlamaGen-XL satisfies neither. 
Because this diagnostic depends only on the upstream model's logit statistics, it provides a practical criterion for predicting when DCDD is likely to help.

\subsection{Transfer to Janus-Pro 1B without Retraining}
\label{sec:janus1b}

Janus-Pro 1B shares the VQ-VAE tokenizer of Janus-Pro 7B, so the
DCDD checkpoint trained for the 7B model can be applied to it
directly. Because the decoder was trained for the 7B conditioning
regime, we align the 1B model's inference-time logits to the 7B
target statistics (Eq.~\eqref{eq:target_stats}) using the same
gradient-free scale, temperature, and label-smoothing search as LC
(Appendix~\ref{sec:extended_lc}), applied at generation time; no
parameters are trained.
Measured with the same protocol as Table~\ref{tab:otsu_comp} (100
GenEval prompts, each model's own CFG and sampling settings, before
top-$k$/top-$p$ filtering), the 1B logits fall in the same regime as
the 7B logits: Otsu rank 2, top-1 probability 0.2457, top-2
probability 0.0978, head mass 0.3435, and normalized entropy 0.4033.

\begin{table}[h]
\centering
\caption{\textbf{Janus-Pro 1B with the Janus-Pro 7B DCDD checkpoint}
(no retraining; mean $\pm$ SD over three seeds). gFID is measured on
the MJHQ-30K Animals category.}
\resizebox{0.6\linewidth}{!}{
\begin{tabular}{lcc}
\Xhline{4\arrayrulewidth}
\rowcolor{table-yellow!60}
\textit{\textbf{Janus-Pro 1B}} & gFID$\downarrow$ (Animals) & GenEval$\uparrow$ \\
\midrule
VQ-VAE Decoder & 35.14 $\pm$ 0.28 & \hl{63.46 $\pm$ 0.28} \\
\textbf{DCDD (ours)} & \hl{26.09 $\pm$ 0.26} & 62.17 $\pm$ 0.45 \\
\Xhline{4\arrayrulewidth}
\end{tabular}
}
\label{tab:janus1b}
\end{table}

Fidelity improves by 25.7\%, while GenEval changes by $-1.29$
consistently across seeds. This matches the intended behavior: DCDD
improves fidelity while following the upstream model's token belief,
so semantic scores are expected to remain close to those of the
native decoder. Within a shared tokenizer, a single DCDD checkpoint
therefore extends across model scales; models with a different
tokenizer require a separately trained decoder.

\section{Qualitative Samples}
\label{sec:qualitative}

\begin{figure}[t]
  \centering
  \includegraphics[width=\linewidth]{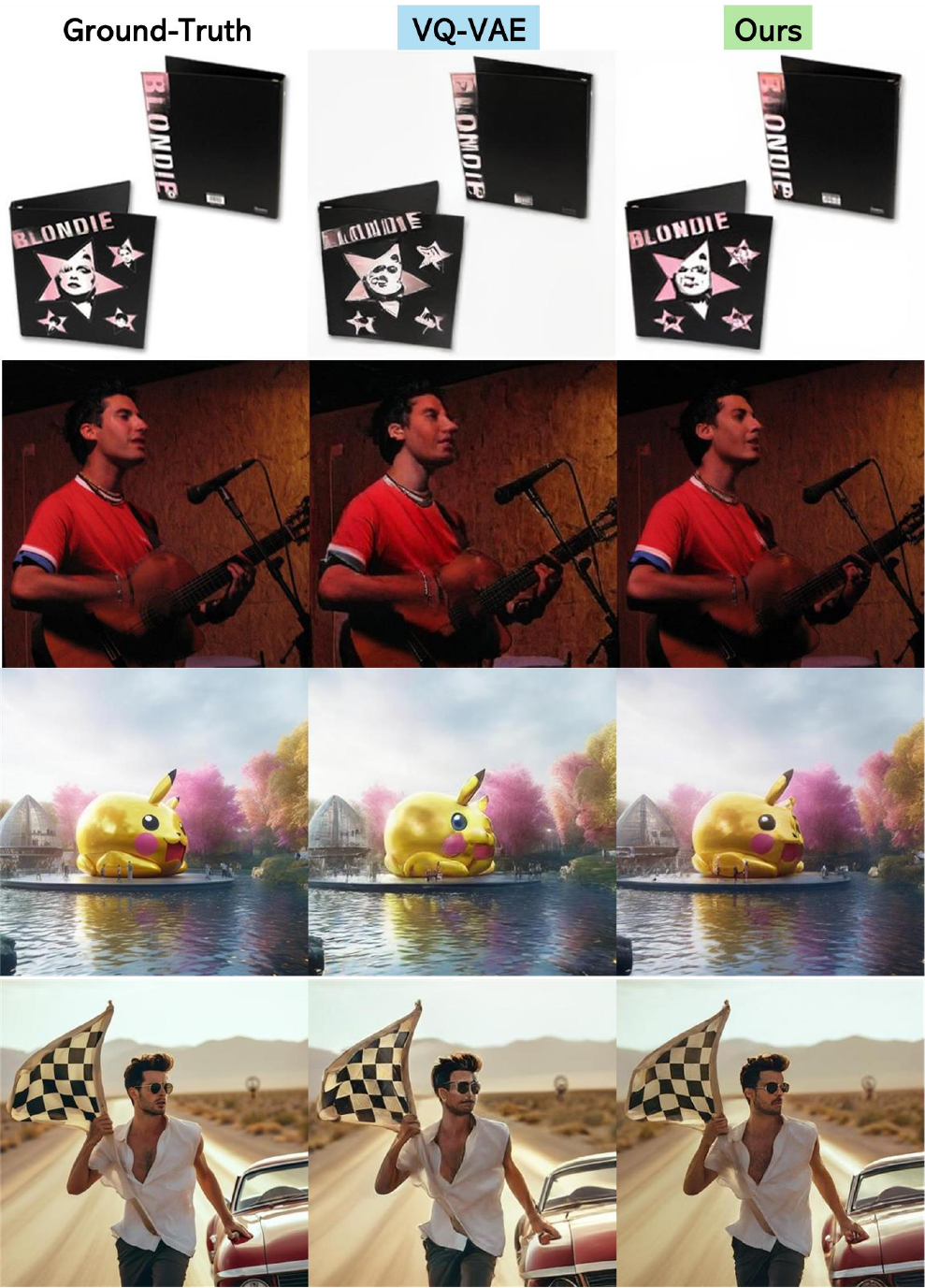}
  \caption{\textbf{Reconstruction samples on Janus-Pro 7B}
  ($384 \times 384$). VQ-VAE decoder versus DCDD on identical
  encoded representations. Top two rows: ImageNet-1K. Bottom two
  rows: MJHQ-30K. DCDD recovers fine structure and texture detail
  that the native VQ-VAE decoder loses.}
  \label{fig:recon_janus}
\end{figure}

\begin{figure}[t]
  \centering
  \includegraphics[width=\linewidth]{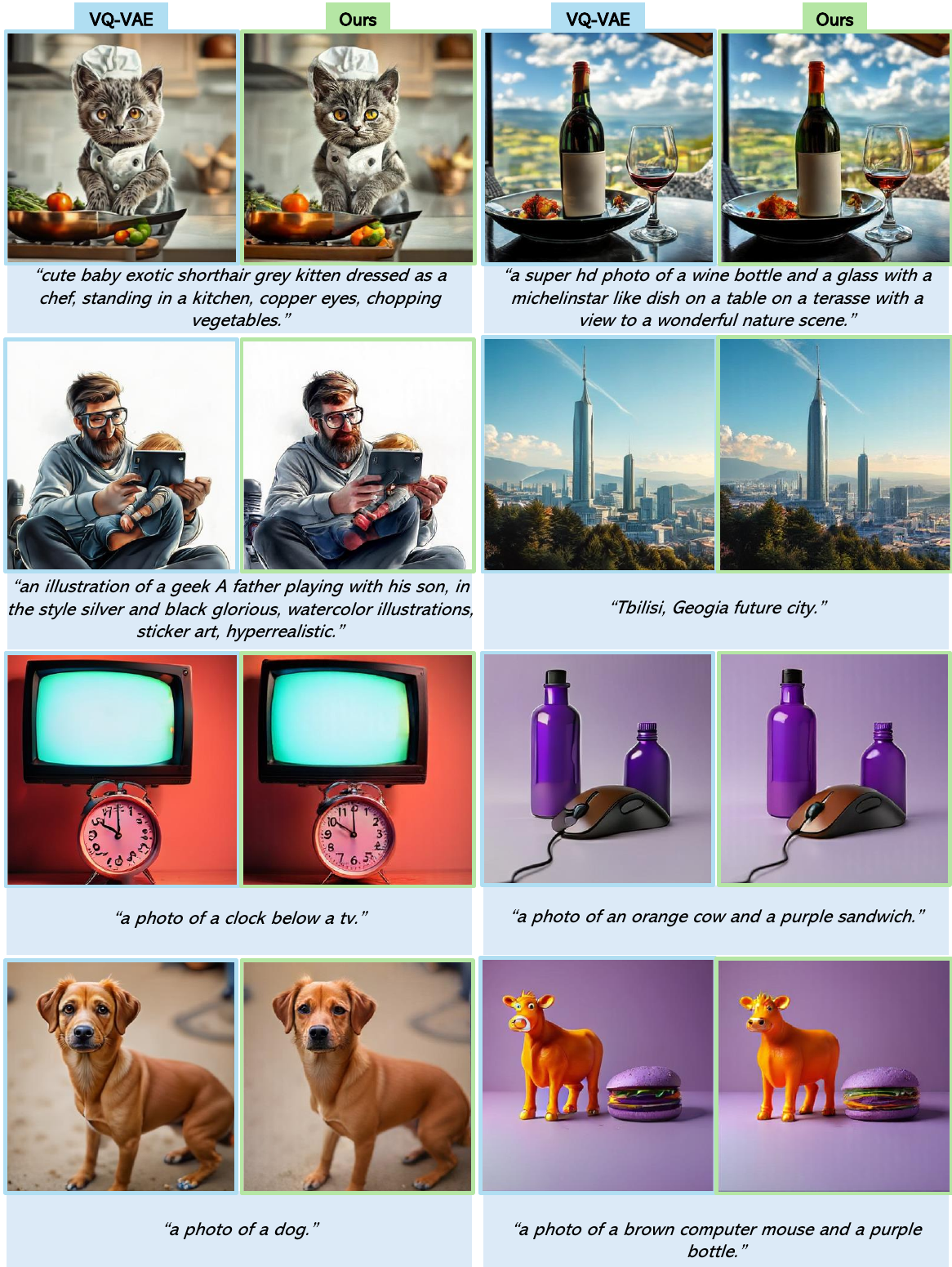}
  \caption{\textbf{Text-to-image samples on Janus-Pro 7B}
  ($384 \times 384$). VQ-VAE decoder versus DCDD on identical
  VLM-predicted token sequences. Top two rows: prompts from
  MJHQ-30K. Bottom two rows: prompts from GenEval. DCDD improves
  visual fidelity while preserving the VLM's semantic intent.}
  \label{fig:t2i_janus}
\end{figure}

\begin{figure}[t]
  \centering
  \includegraphics[width=\linewidth]{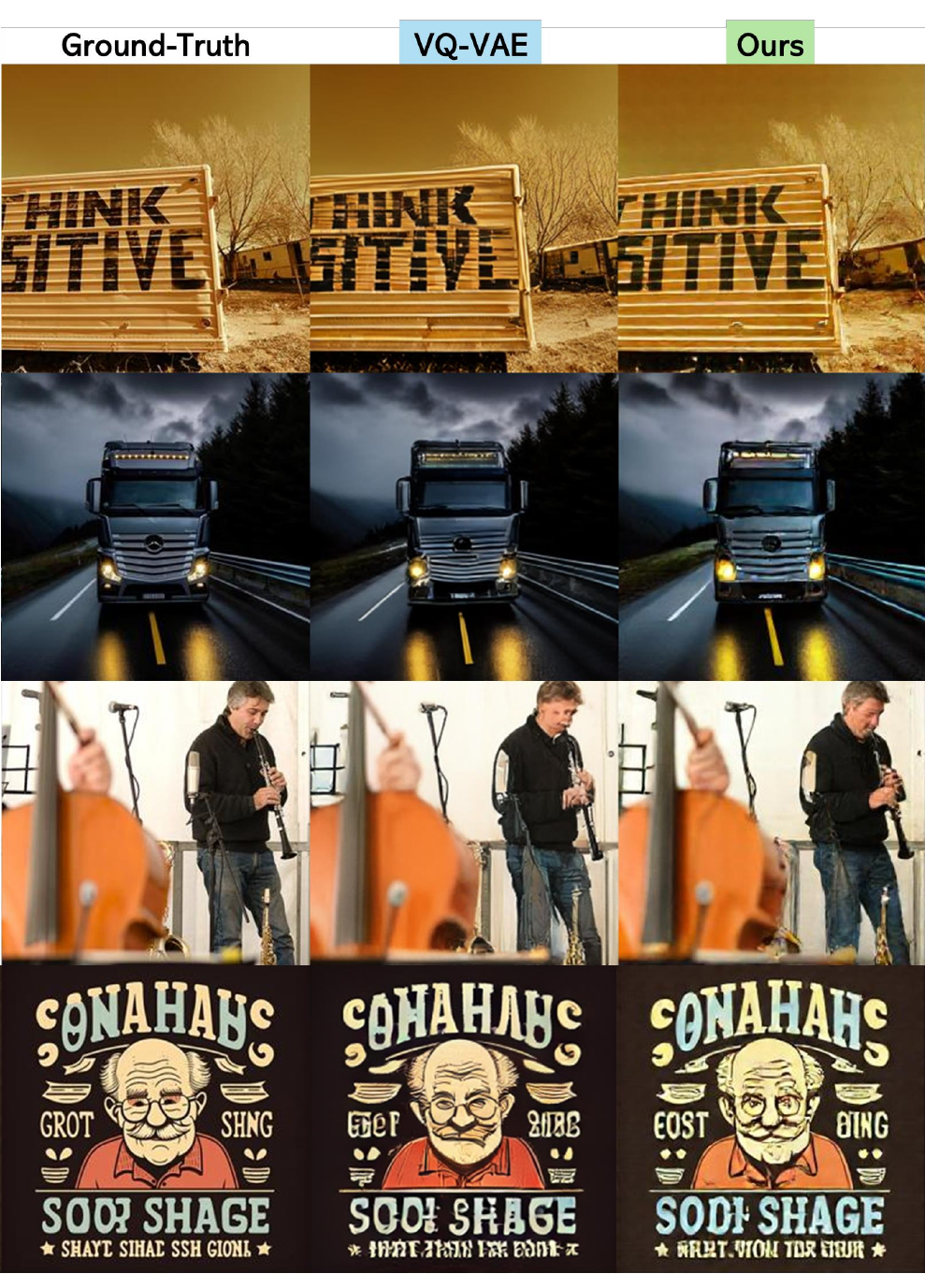}
  \caption{\textbf{Reconstruction samples on LlamaGen-XL}
  ($256 \times 256$). VQ-VAE decoder versus DCDD on identical
  encoded representations. Top two rows: ImageNet-1K. Bottom two
  rows: MJHQ-30K. As with Janus-Pro 7B, reconstruction fidelity
  improves substantially across both datasets.}
  \label{fig:recon_llamagen}
\end{figure}

\begin{figure}[t]
  \centering
  \includegraphics[width=\linewidth]{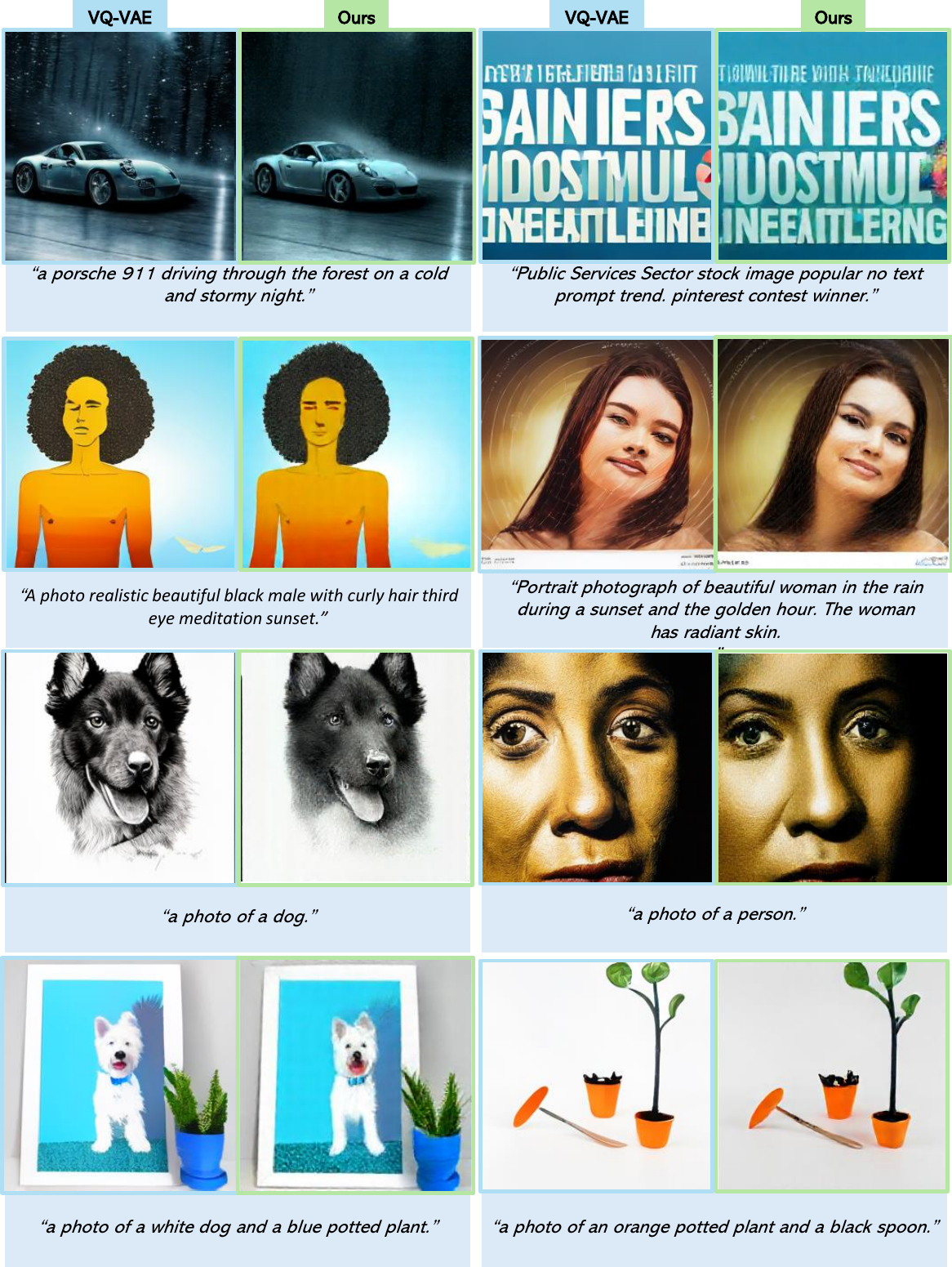}
  \caption{\textbf{Text-to-image samples on LlamaGen-XL}
  ($256 \times 256$). VQ-VAE decoder versus DCDD on identical
  VLM-predicted token sequences. Top two rows: prompts from
  MJHQ-30K. Bottom two rows: prompts from GenEval. Consistent with
  the regime-of-applicability analysis (Appendix~\ref{sec:appendix_vlm}),
  T2I gains are limited on this backbone because the upstream
  token sequence is already semantically degraded.}
  \label{fig:t2i_llamagen}
\end{figure}

This section provides additional qualitative samples for both
backbones evaluated in the paper. Figures~\ref{fig:recon_janus}
and~\ref{fig:t2i_janus} show reconstruction and text-to-image
results on Janus-Pro 7B at $384 \times 384$ resolution;
Figures~\ref{fig:recon_llamagen} and~\ref{fig:t2i_llamagen} show
the corresponding results on LlamaGen-XL at $256 \times 256$. In
each figure, the top two rows draw from ImageNet-1K (reconstruction)
or MJHQ-30K (T2I), and the bottom two rows from MJHQ-30K
(reconstruction) or GenEval (T2I). All comparisons are between the
backbone's native VQ-VAE decoder and DCDD using the same input
tokens.

\section{Limitations}
\label{sec:limitations}

\textbf{Dependence on the upstream VLM.} The effectiveness of DCDD
is shaped by properties of the upstream VLM, as characterized in
Section~\ref{sec:regime} and Appendix~\ref{sec:appendix_vlm}: gains
are most pronounced when the per-token logits retain meaningful
distributional structure and when the baseline token sequence is
already reasonably coherent. Since the decoder is conditioned only
on VLM-derived token representations, it preserves the upstream
model's semantic output rather than correcting it, which is a
property of the post-hoc design choice consistent with our
objective of improving fidelity without altering the VLM's intent.
DCDD is therefore not a remedy for weak upstream generation.

\textbf{Scope across tokenizers.} A decoder is trained for one
tokenizer. Within a shared tokenizer, the same checkpoint transfers
across model scales (Appendix~\ref{sec:janus1b}); models with a
different tokenizer require a separately trained decoder, and we
have not evaluated such models end to end beyond LlamaGen-XL.

\textbf{Distribution shift.} The decoder is trained only on
ImageNet-1K, yet improves MJHQ-30K reconstruction and generation,
and the calibration targets remain stable under unnatural,
compositionally overloaded, self-contradictory, blank, and random
prompts (Appendix~\ref{sec:lc_sensitivity}). We have not, however,
systematically evaluated semantic consistency on long-form or highly
out-of-distribution prompts. The correspondence between training-time
proxies and inference-time VLM logits is established empirically
(Appendix~\ref{sec:proxy_validity}); Appendix~\ref{sec:stability}
bounds the sensitivity of the conditioning but does not provide a
generalization guarantee.

\textbf{Computational cost.} Inference introduces additional latency
from iterative denoising (31\% end-to-end at 30 steps, 10.7\% at 10
steps; Section~\ref{sec:cost}, Appendix~\ref{sec:latency}), and the
full pipeline requires approximately 37.6\,GiB of peak memory, which
exceeds typical edge-device budgets. The native VQ-VAE path remains
available when efficiency is preferred, and a quantized or distilled
decoder is a natural direction for constrained deployment.

\section{Impact Statement}
\label{sec:impact}

Our work improves the visual fidelity of images generated by
pre-trained text-to-image VLMs. As with other advances in image
generation, this could contribute to creating more realistic
synthetic content, with attendant concerns about deceptive use;
these risks are not unique to our method and are addressed by
established safeguards developed for generative AI more broadly.
We note one property specific to our approach: because DCDD does
not introduce its own semantic prior, it will faithfully render
biases or errors present in the underlying VLM. Responsible deployment therefore
requires bias auditing and mitigation at the level of the upstream
VLM rather than at the decoding stage.

\section{Future Directions}
\label{sec:future}

DCDD shows that the pre-sampling distribution of a frozen
discrete-token model is a useful conditioning signal, which suggests
several extensions. On the conditioning side, timestep-adaptive
feature weighting and alignment, effective in diffusion and
flow-matching transformers for virtual try-on and video-to-audio
synthesis~\cite{hong2025itamdt,ton2025taro}, could replace our
uniform injection across denoising steps, and the same principle
applies to speech and audio generators that condition on rich
continuous features~\cite{eom2026transcriptfree}. On the downstream
side, higher-fidelity decoding and per-token uncertainty can benefit
3D scene generation, image-based Gaussian splatting, novel-view
enhancement, and geometry-aware or training-free
editing~\cite{koo2026inspace,lim2026gada,kim2026specsia,koo2025flowdrag,lim2026tango}.
Finally, visual-token distributions could serve as soft targets or
confidence-aware rewards for training smaller or stronger multimodal
models, in line with recent token-level distillation and
confidence-based reinforcement learning for vision-language
models~\cite{jang2026trag,yoon2026decomposed,yoon2026pdcr}.


\end{document}